\documentclass[journal]{IEEEtran}

\usepackage{color}
\usepackage{marginnote}
\makeatletter
\newcommand{\smarginnote}[1]{\begingroup\if@firstcolumn\reversemarginpar\fi\marginnote{#1}\normalmarginpar\endgroup}
\makeatother

\newcommand{\tfflagl}[1]{\if@firstcolumn\normalmarginpar\marginnote{\color{red}\scriptsize{\fbox{#1}}}\else\reversemarginpar\marginnote{\color{red}\scriptsize{\fbox{#1}}}\normalmarginpar\fi}
\newcommand{\tfflagr}[1]{\if@firstcolumn\reversemarginpar\marginnote{\color{red}\scriptsize{\fbox{#1}}}\normalmarginpar\else\normalmarginpar\marginnote{\color{red}\scriptsize{\fbox{#1}}}\fi}

\ifCLASSINFOpdf
\else
\fi
\usepackage{graphicx}
\usepackage{mathtools}
\usepackage{amsmath,amssymb,amsfonts}
\usepackage{algorithm}
\usepackage{algpseudocode}
\usepackage{multirow,multicol}
\usepackage{booktabs}
\usepackage{caption}
\usepackage[mathlines]{lineno}
\usepackage{hyperref}

\usepackage{soul}

\newcommand{\argmin}{\operatornamewithlimits{argmin}}

\begin{document}
%
\title{A Dense Subframe-based SLAM Framework with Side-scan Sonar}
%
%
%

\author{Jun Zhang,
        Yiping Xie,
        Li Ling
        and John Folkesson
\thanks{Jun Zhang is with the Institute of Computer Graphics and Vision (ICG), Graz University of Technology, Austria. Yiping Xie, Li Ling and John Folkesson are with Division of Robotics, Perception and Learning (RPL), KTH Royal Institute of Technology, Sweden.}
}

%
%

\markboth{Preprint}
{Shell \MakeLowercase{\textit{et al.}}: Bare Demo of IEEEtran.cls for IEEE Journals}
%



\maketitle

\begin{abstract}
Side-scan sonar (SSS) is a lightweight acoustic sensor commonly deployed on autonomous underwater vehicles (AUVs) to provide high-resolution seafloor images. However, leveraging side-scan images for simultaneous localization and mapping (SLAM) presents a notable challenge, primarily due to the difficulty of establishing sufficient amount of accurate correspondences between these images. To address this, we introduce a novel subframe-based dense SLAM framework utilizing side-scan sonar data, enabling effective dense matching in overlapping regions of paired side-scan images. With each image being evenly divided into subframes, we propose a robust estimation pipeline to estimate the relative pose between each paired subframes using a good inlier set identified from dense correspondences. These relative poses are then integrated as edge constraints in a factor graph to optimize the AUV pose trajectory.

The proposed framework is evaluated on three real datasets collected by a Hugin AUV. One of these datasets contains manually-annotated keypoint correspondences as ground truth and is used for evaluation of pose trajectory. We also present a feasible way of evaluating mapping quality against multi-beam echosounder (MBES) data without the influence of pose. Experimental results demonstrate that our approach effectively mitigates drift from the dead-reckoning (DR) system and enables quasi-dense bathymetry reconstruction. An open-source implementation of this work is available\footnote[1]{\url{https://github.com/halajun/acoustic_slam}}.
\end{abstract}


%
\IEEEpeerreviewmaketitle

\section{Introduction}

\IEEEPARstart{M}{any} commercial autonomous underwater vehicles (AUVs) rely on the dead-reckoning (DR) system for underwater navigation, which is subject to unbounded errors accumulated from the sensor uncertainties. The common solutions to mitigate such errors involve either integrating global referencing systems (i.e., GPS) that requires periodic resurfacing of the vehicle, or deploying pre-installed acoustic ranging systems such as long/short/ultrashort baseline setups, which could be cumbersome to deploy and confined to limited operational ranges. An alternative solution is to incorporate sensor measurements of the environment to reduce the dead-reckoning drift using a simultaneous localization and mapping (SLAM) method~\cite{thrun2002acm}.

\begin{figure}
 \centering
 \includegraphics[width=1.\columnwidth]{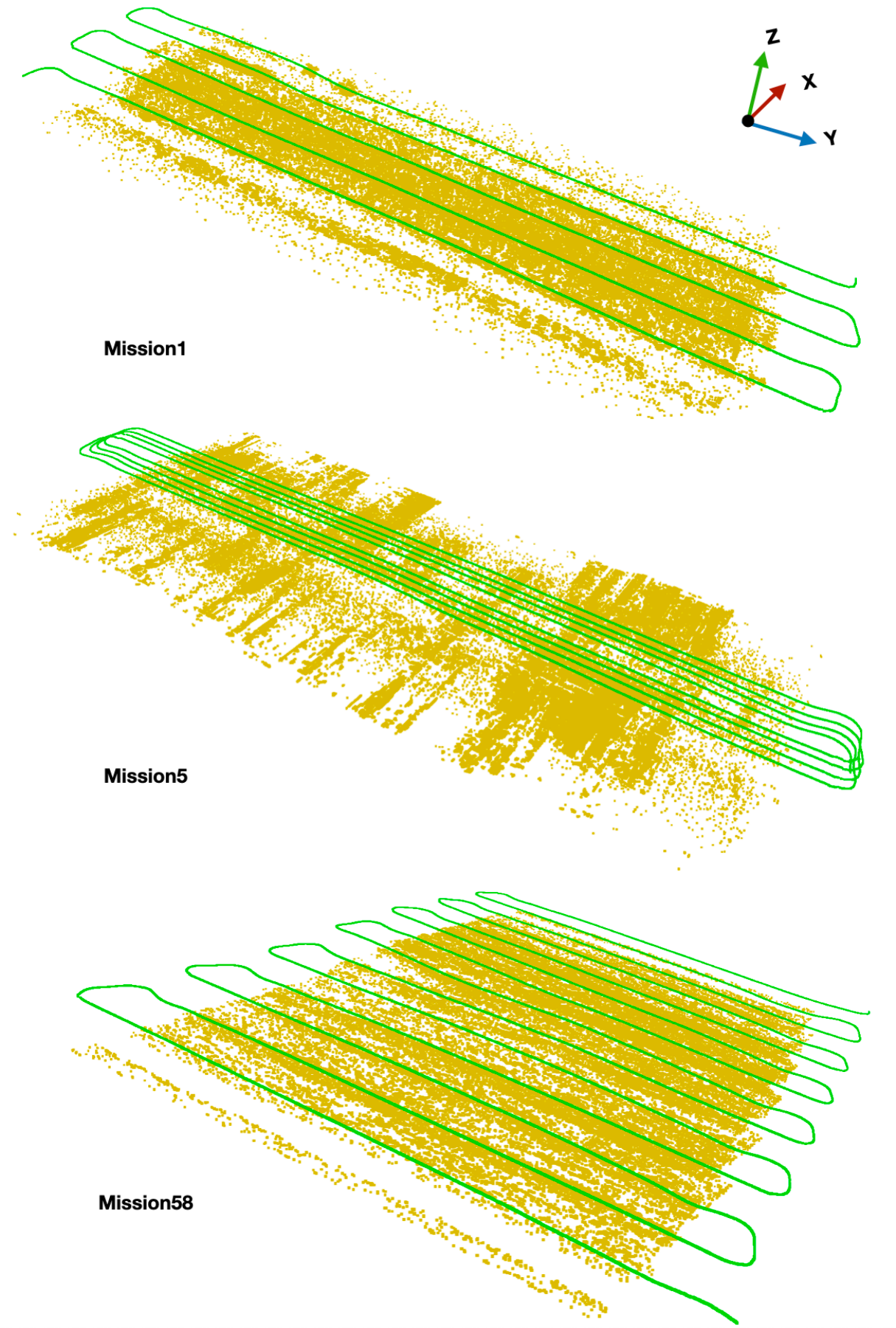}
 \caption{Qualitative results of our proposed dense side-scan SLAM method on the three evaluated mission data named by their mission number. The green lines refer to the estimated trajectories, and the yellow points refer to the reconstructed quasi-dense map.}
\label{fig:showcase}
\end{figure}

For AUVs surveying undersea, three types of sonar sensors are frequently used: side-scan sonar (SSS), useful for producing high-resolution images of a large swath of the seafloor; multi-beam echosounder (MBES), useful for producing the $3$D geometry of the seafloor directly; and forward-looking sonar (FLS), useful for capturing wide-angle, fan-shaped images at consecutive times with large amount of overlap in between. MBES provides $3$D data of the seabed as point clouds that are relatively easy to interpret for underwater SLAM applications~\cite{palomer2016sensors}\cite{torroba2019iros}. The main drawbacks of using MBES for SLAM, however, are its relatively narrow coverage resulting in limited data overlaps between surveying paths, and the difficulty to recognize patterns in terrains that are geometrically featureless, e.g., relatively flat or with gentle slopes. The typical range of FLS is from meters to over a few tens of meters, making it suitable for robust SLAM in small-scale environments~\cite{westman2018icra}\cite{li2018ral}\cite{westman2019joe}. Nevertheless, the utilization of forward-looking sonar comes with unique challenges, including low image texture and high signal-to-noise ratio (SNR). In contrast, SSS is generally used to map seafloor of hundreds and
thousands of meters in large-scale environments. 
Compared to MBES, SSS offers higher resolution that allows for precise AUV localization, as well as larger coverage that allows for localization and mapping using large areas of the seafloor.

However, the problem of SLAM with SSS images is far from being solved, mainly due to the challenge of registering SSS images to one another and the lack of $3$D information. In particular, raw SSS images are geometrically distorted and unevenly ensonified. 
As a result, the appearance of same region of seabed can vary in SSS images when observed from different positions. 
To address this, we first apply a canonical transformation to the raw SSS images to reduce such distortions and redistribute the image pixels to be approximately equal size patches of the seabed. Then, we propose an effective dense (pixel-wise) matching method based on~\cite{barnes2009tog}, combining both geometric and appearance constraints of side-scan image, to find dense correspondences between overlapping images. 

The inability of measuring $3$D data makes the estimate of AUV pose underdetermined using single-ping measurement, even at the presence of accurately associated correspondence~\cite{zhang2023ietrsn}. To mitigate such underdetermination, we propose to evenly divide each SSS image into subframes and estimate the relative pose of the centre pings between paired subframes using multiple ping measurements from dense correspondences as constraints. To be robust against noise and outliers in the dense correspondences, we integrate the pose estimation into a RanSaC (Random Sample Consensus)~\cite{fischler1981acm} pipeline, using both measurement cost and optimization cost as iteration signal.  




With the above considerations, we propose to estimate the AUV poses and landmarks of paired correspondences as a subframe-based graph SLAM problem solved in two steps. First, we model the selected pixel correspondences in each pair of associated subframes as $2$D measurements. These measurements are formulated together with dead-reckoning constraints as a least-squares minimization problem in a robust estimation framework, to deliver an accurate relative pose constraint between the two associated subframes. In the second step, all of the estimated poses are considered as loop-closure constraints in a pose graph for global optimization that refines the entire AUV pose trajectory. Compared to sparse ping/feature-based approaches~\cite{zhang2023ietrsn}\cite{aulinas2010oceans}, our proposed method is able to perform more accurate AUV pose estimation, as well as generate quasi-dense bathymetry point cloud using the dense correspondences and optimized pose trajectory. 

We demonstrate the performance of our proposed method using carefully-considered metrics on three real datasets from different surveying missions. This includes evaluations on pose accuracy by hand-annotating a set of images as ground truth reference. For mapping evaluation, we propose to align the reconstructed point cloud of each SSS image to the MBES point cloud of the same surveying line, and compare their heightmaps, which can avoid the influence of pose trajectory being inaccurate. In summary, our contributions are:

\begin{itemize}
    \item We propose an effective dense matching method for side-scan image, and utilize it to achieve robust and accurate pose estimation between side-scan subframes;
    \item We introduce a subframe-based dense side-scan sonar SLAM framework that is able to refine the AUV pose trajectory from dead-reckoning system, as well as reconstruct a quasi-dense bathymetry of the seafloor, and open source it for the benefit of the community;
    \item We present a feasible way of evaluating the performance of underwater SLAM methods, which is very challenging for  underwater scenarios. This is done using manually-annotated keypoints and MBES data collected together.
\end{itemize}

In the following, we show in details the methodology and implementation of our proposed framework in Sec.~\ref{sec:method}, after related work in Sec.~\ref{sec:related_work}. Experimental results are documented in Sec.~\ref{sec:experi}, with concluding remarks summarised in Sec.~\ref{sec:concl}.


\section{Related Work}
\label{sec:related_work}

\subsection{SLAM with Side-scan Sonar}

Research on SLAM with SSS has drawn considerable attention within the community in the past two decades. Among the earlier works, \cite{ruiz2003oceans}\cite{ruiz2004joe}\cite{reed2006tip} combine a stochastic map with Rauch-Tung-Striebel (RTS) filter to estimate AUV's location, where the stocahstic map is formulated as extended Kalman filter (EKF), with landmarks manually extracted from the side-scan images. Later in~\cite{fallon2011icra}, the authors propose to fuse acoustic ranging and side-scan sonar measurements in pose graph SLAM state estimation framework that avoids inconsistent solution caused by information lost during linearization~\cite{julier2001icra}, to achieve accurate AUV trajectory estimation. Similarly, \cite{bernicola2014oceans} proves the capabilities of using SSS images in graph-based iSAM~\cite{kaess2008tro} algorithm to produce corrected sensor trajectories for image mosaicing. Issartel et al.~\cite{issartel2017oceans} take a further step to incorporate switchable observation constraints in a factor graph to address false data association issue and guarantee robust solution. The above solutions prove the feasibility of using SSS information to help better localize AUVs. However, they rely on manual annotation of landmarks in SSS images, which goes against full autonomy in AUVs.

There are but few approaches attempting to concurrently address the problems of both SSS image association and SLAM. \cite{aulinas2010oceans}\cite{aulinas2010iros} proposes to use a boosted cascade of Haar-like features to perform automatic feature detection in SSS images. The correspondences are found by means of joint compatibility branch and bound (JCBB) algorithm~\cite{neira2001tra}, and used to update pose states in an EKF SLAM framework. By applying a similar framework, \cite{siantidis2016auv} combines local thresholds and template-based detectors with strict heuristics to avoid false associations. The integration of JCBB in EKF is arguably well-fitted to solve the complete SSS SLAM problem with few landmark measurements, but it suffers from inconsistent performance and offers lower accuracy when comparing to graph-based SLAM solution~\cite{fallon2010ijrr}.

\subsection{Feature Detection and Matching of Side-scan Image}

Some efforts have been made to exploit feature extraction and matching in side-scan images, mainly for estimating the relative pose of AUV through image registration. \cite{vandrish2011sutwsuscrt} is one of the earliest works that try to apply feature-based registration method using SIFT~\cite{lowe2004ijcv} for side-scan sonar images. Later in~\cite{king2013oceans}, the authors extensively compare the performance of off-the-shelf feature detection techniques in matching side-scan sonar image tiles, and integrate some of the robust solutions into an image registration framework for their AUV route following system~\cite{king2012auv}. Lately, the fusing of visual and geometric information to accomplish data association has become a trending solution. For instance, \cite{mackenzie2015ccece} uses elevation gradients as additional features to assist with associations between landmarks. In~\cite{petrich2018oe}, the authors propose to construct features with $3$D location and feature intensity, and associate between them through a modified version of iterative closest point (ICP)~\cite{besl1992sf} scan matching method. While interesting conclusions are drawn in the above works, none has integrated an applicable solution into a full SLAM framework.

\begin{figure}
 \centering
 \includegraphics[width=1.0\columnwidth]{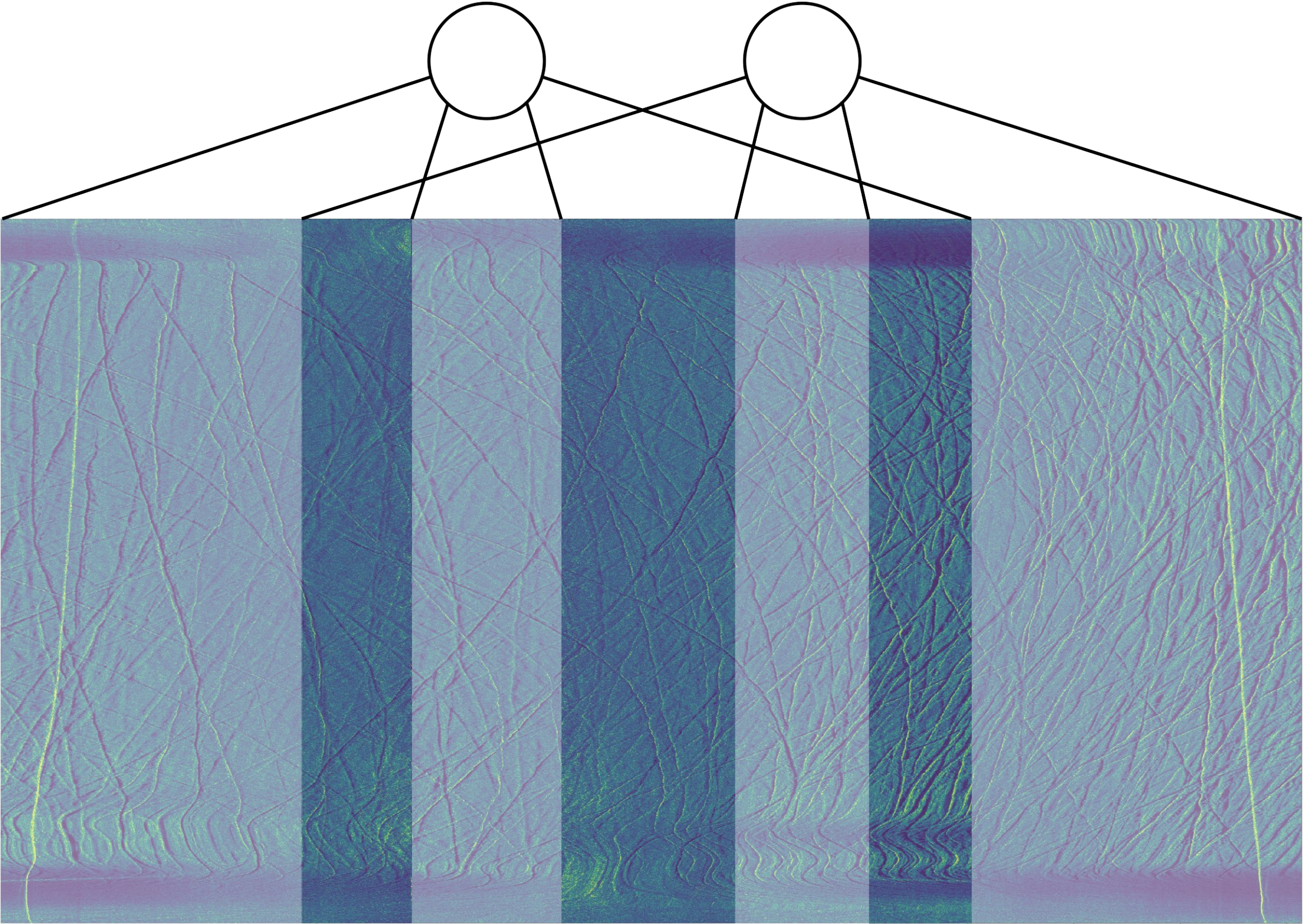}
 \caption{A demonstration of overlapping side-scan images captured along two adjacent survey lines from our self-collected data (Mission58), where the sketch on top represents the AUV and its side-scan covering ranges. The non-transparent areas of the bottom images indicate the overlapped parts.}
 \label{fig:ss_demo}
\end{figure}

\subsection{Dense Matching of Optical Image}

Large displacement optic-flow estimation is a task to establish pixel-wise correspondences between optical images with large parallax, and finding dense correspondences between parallel side-scan images is closely related to this task. While classic variational approaches~\cite{brox2010pami}\cite{weinzaepfel2013iccv} have been tried by integrating sparse descriptor matching in a cross-to-fine scheme to mitigate the large displacement issue, they typically assume local smoothness and that motion discontinuities only happen at the image edges~\cite{revaud2015cvpr}. These assumptions do not apply to side-scan images, since there are discontinuities at the nadir of each side-scan image and when the trajectories of forming parallel side-scan images are close to each other, see Fig.~\ref{fig:ss_demo}. 

In contrast, methods~\cite{he2012cvpr}\cite{korman2015pami}\cite{bailer2015iccv}\cite{hu2016cvpr} based on nearest neighbor fields (NNF) are more suitable for our problem, as they are able to perform efficient global search for best match directly on full image resolution. Nevertheless, NNF-based methods usually generate many outliers that are difficult to identify, due to lack of regularization. In this paper, we propose an effective dense matching method for side-scan image, by carefully modifying on a NNF-based seminal work called PatchMatch~\cite{barnes2009tog} to combine both dead-reckoning and imaging information to search for dense correspondences. To mitigate the influence of outliers to pose estimation, a robust RanSaC-based pose estimation pipeline is introduced. 


\section{Methodology}
\label{sec:method}

\begin{figure*}
 \centering
 \includegraphics[width=2.\columnwidth]{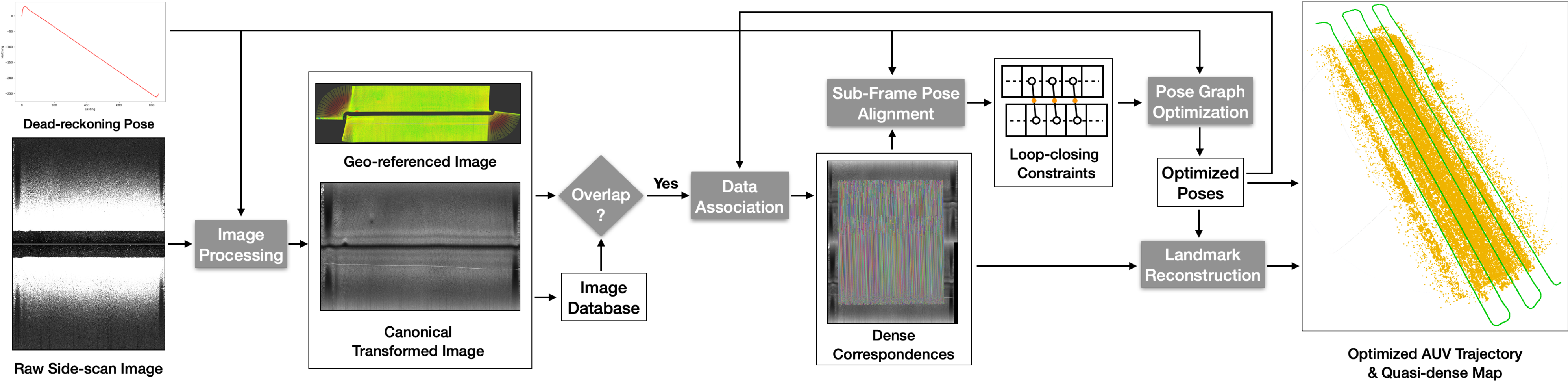}
 \caption{Overview of our subframe-based side-scan sonar SLAM framework. The framework takes raw side-scan images with dead-reckoning poses as input and outputs an optimized pose trajectory together with a quasi-dense map. The grey color rectangles highlights the main components, while rectangles with no color-filled represents intermediate outputs.}
 \label{fig:framework}
\end{figure*}

In this work, assuming that the AUV follows a common lawnmower or loop pattern with major overlap between survey lines, the proposed framework considers the side-scan image formed by each line as input. Note that, each raw SSS ping contain over $40k$ bins, and this high across-track resolution, together with the low along-track resolution ($\sim$$0.5$m per ping), significantly distorts the seabed appearance. To avoid such issue, the SSS pings are down-sampled to $1$$\sim$$3k$ bins (varying by datasets) per side (port and starboard), which is sufficient to capture the detailed texture information of the seabed.

In the following, we describe the details of our dense SSS SLAM framework, which takes the down-sampled sonar image and dead-reckoning data as input, and outputs an optimized AUV pose trajectory and a quasi-dense map. The proposed pipeline is summarized in Fig.~\ref{fig:framework}, and it contains five parts: image processing, data association, subframe pose estimation, pose graph optimization and landmark reconstruction.

\subsection{Image Processing}

\subsubsection{Canonical Transformation}

Data association across different SSS images is challenging, as the same physical area on the seabed will appear differently in the SSS images collected from different distances and angles. Furthermore, an SSS image will be distorted compared to an orthographic projection of the ensonified region and will vary in intensity as a function of the sonar position. To reduce such distortions, we apply a method~\cite{xu2023oceans} capable of transforming SSS images from different survey lines into a canonical representation, which includes two steps: intensity correction and sensor-independent slant range correction. 

Intensity correction is performed to normalize the intensities across images based on the ``Lambertian'' model~\cite{coiras2007tip}\cite{burguera2014etfa}, with the SSS backscatter intensity being proportional to the cosine square~\cite{aykin2013oceans} of the incidence angle. The slant range correction adjusts for the varying projection of side-scan pixels on the assumed horizontal seafloor. An adjustment of sine of the incidence angle is performed so that the pixels can be of fixed size in terms of horizontal range rather than slant range. More details can be found in~\cite{xu2023oceans}. 

\subsubsection{Geo-referenced Image}

To approximately check whether any two of the SSS images are overlapping, we follow the method in ~\cite{king2012auv} and utilize the dead-reckoning data to geo-reference each pixel in the SSS images. This provides an approximate location of each pixel of an SSS image in global reference coordinates. These geo-references are also used to narrow the searching area when performing dense matching in data association. 

\subsection{Data Association}
\label{sec:da}

Once an overlapping area is found between a new SSS image and any image in the database, a data association process is performed to find dense pixel correspondences between the overlapping images. In this work, we propose to accomplish this by formulating it as a $\it{nearest}$-$\it{neighbor}$ $\it{field}$ (NNF) estimation problem, and solving it with computing the approximate NNF based on our modified version of PatchMatch~\cite{barnes2009tog}, which is carefully tailored for side-scan images. 

\begin{figure}
 \centering
 \includegraphics[width=1.\columnwidth]{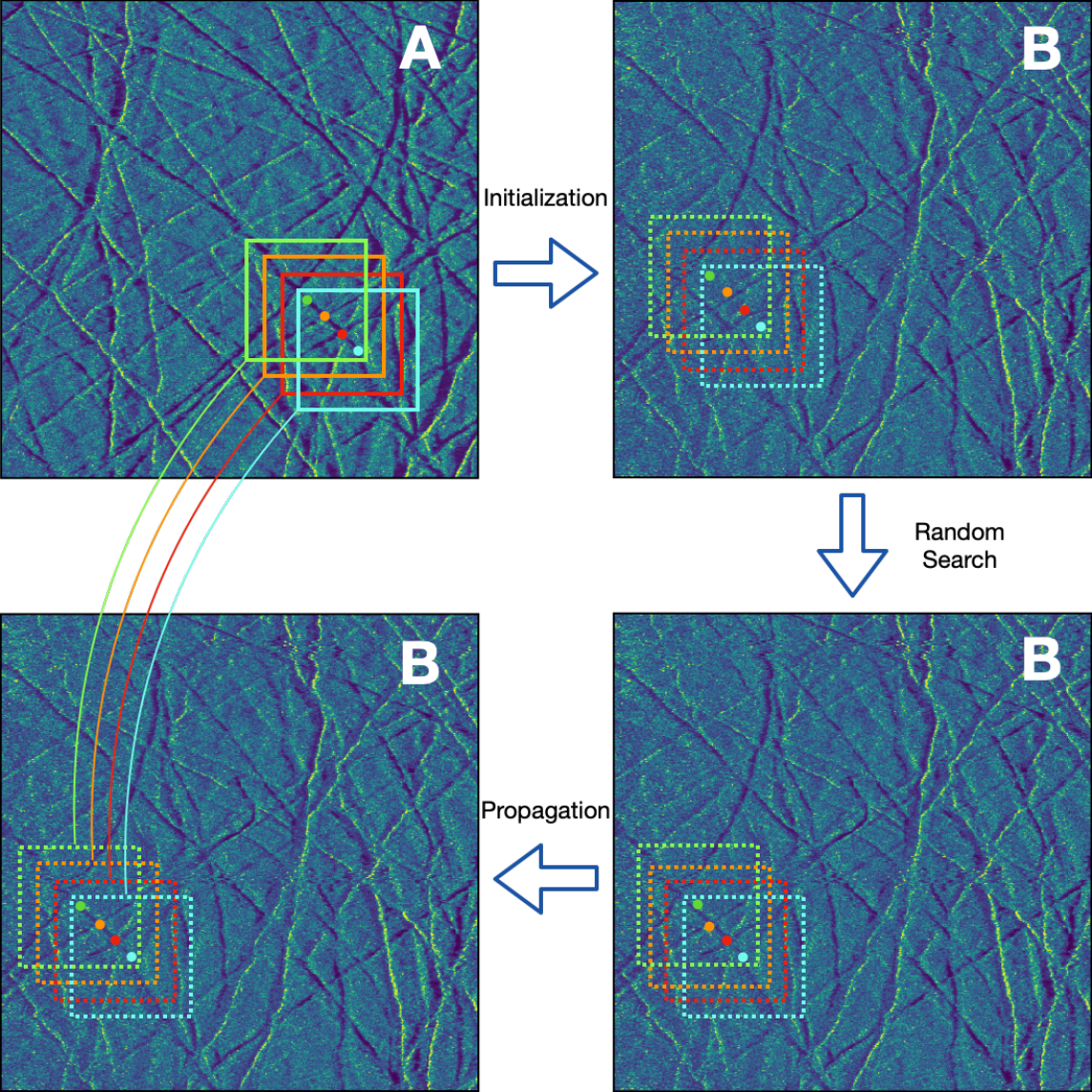}
 \caption{
 Illustration of the 3-step dense matching algorithm using two overlapping SSS images ($\mathbf{I}_{A}$ and $\mathbf{I}_{B}$). The figure demonstrates how the 4 sample patches in $\mathbf{I}_{A}$ (shown as rectangles with centre points in different colors) are aligned with areas in $\mathbf{I}_{B}$.
 After {\it{initialization}}, the approximate correspondences of the 4 patches (shown as dashed rectangles in corresponding colors) are found in $\mathbf{I}_{B}$ up to DR precision (top right). The positions of these correspondences are then adjusted through {\it{random search}} (bottom right). Finally, the {\it{propagation}} step further refines the results, leading to smooth and accurate correspondences (bottom left). Note that the last two steps are usually repeated a few times to obtain a satisfactory result.
}
 \label{fig:densematch}
 \vspace{-.4cm}
\end{figure}

Specifically, define a nearest-neighbor field $\mathcal{F}:\mathbf{I}_{A}\mapsto \mathbb{R}^2$ over all possible patch coordinates (locations of patch centers) in a side-scan image $\mathbf{I}_{A}$, so that given a patch coordinate $\mathbf{a}$ in image $\mathbf{I}_{A}$, its corresponding nearest neighbor $\mathbf{b}$ in an overlapping side-scan image $\mathbf{I}_{B}$ can be denoted as,
\begin{eqnarray}
\mathcal{F}(\mathbf{a}) = \mathbf{b} - \mathbf{a}.
\label{eq:nnf}
\end{eqnarray}
The values of $\mathcal{F}$ are referred to as $\it{offsets}$ and stored as a matrix with the same dimension of the image size of $\mathbf{I}_{A}$. Following the pipeline in~\cite{barnes2009tog}, the estimation of $\mathcal{F}$ has three main steps: initialization, random search and propagation, see Fig.~\ref{fig:densematch} and Algorithm~\ref{alg:dm}.

\begin{algorithm}
\caption{Dense Matching for Side-scan Images}\label{alg:dm}
\begin{algorithmic}[1]
\Require \\ $\mathbf{I}_{A},\mathbf{I}_{B},\mathbf{G}_{A},\mathbf{G}_{B}$: side-scan, geo-referenced images; \\
$p_{size},n_{max},o_{max}$: patch size, max iterations, max offset;
\Ensure $\mathcal{F}$: nearest-neighbor field between $\mathbf{I}_{A}$ and $\mathbf{I}_{B}$;
\State $\mathcal{F},\mathcal{D} \gets Initialization(\mathbf{I}_{A},\mathbf{I}_{B},\mathbf{G}_{A},\mathbf{G}_{B})$
\While{$n < n_{max}$}
\State $\mathcal{F},\mathcal{D} \gets RandomSearch(\mathcal{F},\mathcal{D},\mathbf{I}_{A},\mathbf{I}_{B},{p}_{size},{o}_{max})$
\State $\mathcal{F},\mathcal{D} \gets Propagation(\mathcal{F},\mathcal{D},\mathbf{I}_{A},\mathbf{I}_{B},{p}_{size},{o}_{max})$
\State $n$++
\EndWhile
\State \Return $\mathcal{F}$
\end{algorithmic}
\end{algorithm}

\subsubsection{Initialization}\label{sec:ini}
Instead of assigning random values by independent uniform sampling across the full range of image $\mathbf{I}_{B}$ as PatchMatch does, we propose to utilize the geo-referenced image obtained from dead-reckoning data to find the nearest correspondence by comparing Euclidean distance in geometric space. By doing so, we restrict the search space to dead-reckoning error range, which requires less iteration steps, and helps to improve the robustness of finding accurate correspondences. To accelerate the search process of nearest correspondence, we build a kd-tree~\cite{friedman1977toms} to store geo-referenced data and query effectively. Here, the $\it{patch}$ $\it{distance}$ $\it{cost}$ $\mathcal{D}:\mathbf{I}_{A}\mapsto \mathbb{R}$ corresponding to $\mathcal{F}$ is also recorded for iteration, and is initialized with infinity values.

\subsubsection{Random Search}
Based on the initialized offset at each position, a random search is performed around it within a $\it{max}$ $\it{offset}$ ${o}_{max}\in \mathbb{Z}^{+}$, which can be determined by prior knowledge of navigation drift and imaging resolution. Then, given a randomly selected position, we compare the patch distance against its current position. If the randomly selected position yields a smaller patch distance than the current position, then the offset is updated to represent the newly selected position.
The distance function we use in this paper is Zero-mean Normalized Cross-Correlation (ZNCC) distance, which is invariant to linear brightness and contrast variations that often appear close to the nadir and the sides in side-scan images.

\subsubsection{Propagation}\label{sec:ppg}
Assuming that the patch offsets are likely to be the same locally, the propagation step aims to improve the values of $\mathcal{F}$ using the known surrounding offsets. Specifically, let $\mathcal{D}(\mathbf{a})$ be the patch distance between the patch at $\mathbf{a} = (x,y)$ in image $\mathbf{I}_{A}$ and the patch at $\mathbf{b} = \mathbf{a} + \mathcal{F}(\mathbf{a})$ in $\mathbf{I}_{B}$, we take the new value for $\mathcal{F}(\mathbf{a})$ to be the minimum patch distance within the neighbors:
\begin{equation}
\begin{aligned}
\mathcal{F}(\mathbf{a^{*}}) = &\argmin_{\mathbf{a}} \{\mathcal{D}(\mathbf{a} = (x+n,y+n))\},  \\
& \textrm{s.t.} \quad n\in \mathbb{Z} \quad\& \quad n\subseteq  \left [-1,1  \right ].
\label{eq:ppg}
\end{aligned}
\end{equation}
The effect is that if $\mathbf{a} = (x,y)$ has a correct mapping and is in a coherent region, then all neighbors around $\mathbf{a}$ will be filled with the correct mapping. 
However, side-scan images exhibit discontinuities due to multiple reasons, including canonical transformation induced centre-column discontinuity, noisy areas induced by AUV turning motion, as well as sensor artifacts. These continuities are challenging for the propagation. 
To address this, we mask out these areas and exclude them from the matching process.

\subsubsection{Improving Data Association using Optimized Poses}~\label{sec:iter}

The presets of patch size $p_{size}$ and max offset $o_{max}$ cannot be too high, as that would in general lower the patch matching accuracy, which makes it difficult to handle large offsets/drifts in cases where the paired images are distant from each other. To address this, we propose to improve the dense matching accuracy in distant paired images with optimized poses in an iterative fashion (see Fig.~\ref{fig:framework}), which in turn helps to improve the pose accuracy. Here, a new parameter $n_{iter}$ representing the iteration number of this process is introduced. Detailed discussion can be found in Section~\ref{sec:dis2}.

\subsection{Subframe Pose Estimation}\label{sec:spe}

Now we show how each pair of the pixel correspondences is used to construct a measurement model. Based on that, we propose a robust pose estimation pipeline, which is able to estimate the relative pose between a pair of subframes of SSS images, by iteratively identifying a set of good quality inliers from dense correspondences using RanSaC. In the following, we assume that each side-scan image has been evenly divided into multiple subframes and they are used as unit image for estimation instead of the whole one, see Fig.~\ref{fig:factorgraph} left.

\subsubsection{Side-scan Measurement Model}
We model the $i^{th}$ measurement of a bin (pixel) found in a side-scan image as a $2$D measurement that constrains the slant range to the bin, and that it lays in a plane perpendicular to the sonar array:
\begin{eqnarray}
{\mathbf{z}}_{i} = \begin{pmatrix} r_i \\ 0\\ \end{pmatrix} = \hat{\mathbf{z}}_{i} + {\mathbf{\eta}}  = \begin{pmatrix} \sqrt{{\mathbf{\pi}}({}^{o}{\mathbf{x}})\cdot{\mathbf{\pi}}({}^{o}{\mathbf{x}})}  \\ (1,0,0)\cdot{\mathbf{\pi}}({}^{o}{\mathbf{x}})\\ \end{pmatrix} + {\mathbf{\eta}},
\label{eq:KMM}
\end{eqnarray}
\noindent where ${\mathbf{\eta}}$ is measurement noise, ${}^{o}{\mathbf{x}}\in {\rm I\!R}^3$ is the $3$D landmark on the seafloor in a global frame coordinate $o$ that is observed from the $i^{th}$ measured bin, and $r_{i}$ is the range to this landmark. ${\mathbf{\pi}}(\cdot)$ is a function that transforms a $3$D landmark from the global frame coordinate to the sensor frame coordinate $s$:
\begin{eqnarray}
{}^{s}\bar{\mathbf{x}} = {}^{b}{\mathbf{T}}_{s}^{-1}\cdot{}^{c}_{}{\mathbf{T}}_{b}^{-1}\cdot{}^{o}_{}{\mathbf{T}}_{c}^{-1}\cdot{}^{o}\bar{\mathbf{x}}.
\label{eq:transform}
\end{eqnarray}
\noindent Here ${}^{o}_{}{\mathbf{T}}_{c}\in \mathrm{SE}(3)$ is the centre ping pose\footnote{we omit the subscript '$c$' later to avoid ambiguity with ping index.} of the subframe that contains the measured bin. ${}^{c}{\mathbf{T}}_{b}\in \mathrm{SE}(3)$ is the transformation from centre ping to the ping of the measured bin, and we assume that the drift between them is negligible given the size of a subframe being set to reasonably small. ${}^{b}{\mathbf{T}}_{s}\in \mathrm{SE}(3)$ is sensor offset from the ping with measured ping to the sensor, which is normally assumed as fixed and known. Note that ${}^{o}\bar{\mathbf{x}}\in {\rm I\!E}^3$ denotes the homogeneous representation of ${}^{o}{\mathbf{x}}$.

\subsubsection{Relative Pose Estimation}

Given a set of $N$ pixel correspondences between two overlapped subframes $\mathbf{I}_{1}$ and $\mathbf{I}_{2}$, the centre ping poses of both subframes ${}^{o}\mathbf{T}_{1}$ and ${}^{o}\mathbf{T}_{2}$, and the observed $3$D landmarks ${}^{o}{\mathbf{x}_{1:N}}$=$\{{}^{o}{\mathbf{x}_{i}}\,|\,i = 1,...,N\}$ can be estimated via minimizing the minus log of least squares cost, assuming the measurement noises as Gaussian: 
\begin{eqnarray}
\begin{aligned}
& \mathcal{C}({}^{o}{\mathbf{T}_{1}},{}^{o}{\mathbf{T}_{2}},{}^{o}{\mathbf{x}_{1:N}})= 
\frac{1}{2} (\sum^{2N}_{i=1}[(\hat{\mathbf{z}}_i-{\mathbf{z}}_i)\Sigma{}^{-1}_{i}(\hat{\mathbf{z} }_{i} -{\mathbf{z} }_{i}){}^{\top}] \\
& + [({}^{1}\hat{\mathbf{T}}_2-{}^{1}{\mathbf{T}}_2)\Sigma{}^{-1}_{t}({}^{1}\hat{\mathbf{T}}_{2} -{}^{1}{\mathbf{T}}_{2}){}^{\top}]) + \phi({}^{o}{\mathbf{T}}_{1}),
\end{aligned}
\label{eq:costfunc}
\end{eqnarray}
\noindent with the first term on the right side being the side-scan measurement cost calculated by Eq.~\ref{eq:KMM}, the second term the odometry measurement cost and the third term a prior cost. Specifically, ${}^{1}\hat{\mathbf{T}}_{2} = {}^{o}{\mathbf{T}}_{1}^{-1}\cdot{}^{o}{\mathbf{T}}_{2}$ is the relative pose between the centre pings of the subframes, and ${}^{1}{\mathbf{T}}_2$ is the measurement obtained from dead-reckoning data. The odometry term is crucial as it helps to relieve the underconstrained issue in side-scan measurement and ensure convergence to a desired minimum. $\phi(\cdot)$ is a prior model that sets one of the poses (e.g., ${}^{o}{\mathbf{T}}_1$ here) fixed and only adjust for the other, which is reasonable as we aims to estimate the relative pose between them rather than estimate both.

\begin{figure}
 \centering
 \includegraphics[width=1.\columnwidth]{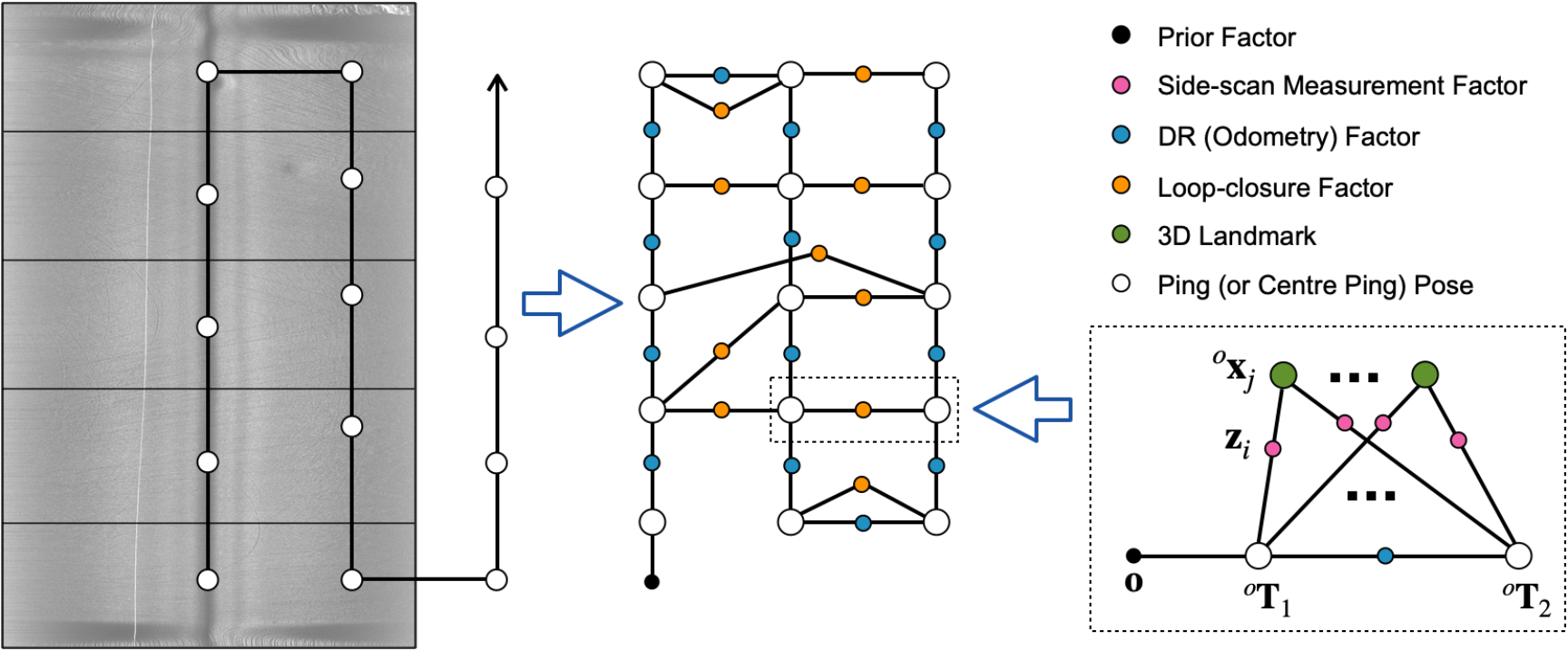}
 \caption{Factor graph representation of our proposed SSS SLAM framework. Left: an AUV surveying trajectory and a demo of an evenly-divided side-scan image aligned to the first survey line. Centre: factor graph of global pose graph optimization. Right: notations (top) and factor graph of relative pose estimation (bottom).}
 \label{fig:factorgraph}
 \vspace{-.4cm}
\end{figure}

$\Sigma{}_{i}$ is the covariance matrix for each side-scan bin measurement defined in Eq.~\ref{eq:KMM}. The variance in the range measurement $r_i$ is dominated by the discretization of the range in the side-scan, varying from $0.01$-$1m^{2}$ depending on the resolution of the image. For the second row restricting the landmark to lie in the plane, the variance should grow as the square of the horizontal distance from the sonar. Since the sonar is mostly used at shallow grazing angles, we can instead use the range so that:
\begin{eqnarray}
\Sigma{}_{i} = \left(\begin{matrix} \sigma^{2}_{r} &0 \\ 0 &r^{2}_{i}\alpha^{2} \end{matrix} \right),
\label{eq:covariance}
\end{eqnarray}
\noindent where $\alpha$ is the beam width in radians. $\Sigma{}_{t}$ is the odometry covariance and is set proportional to the distance between the two dead-reckoning poses. A factor graph describing the costs in Eq.~\ref{eq:costfunc} as factors, is illustrated in Fig.~\ref{fig:factorgraph} bottom right (dashed rectangle).

Eq.~\ref{eq:costfunc} can be solved iteratively using an optimization method such as Levenberg-Marquardt algorithm. However, the dense correspondences between each pair of subframes are large in quantity ($1$-$30$k), and they could be quite noisy and contain lots of outliers, hence it is not suitable to use them all for the optimization. To robustly identify the potential inliers to achieve accurate results, we propose to integrate the optimization into a RanSaC-based estimation pipeline, see Algorithm~\ref{alg:ransac}. 

\begin{algorithm}
\caption{Robust Subframe Pose estimation}\label{alg:ransac}
\begin{algorithmic}[1]
\Require: \\
${}^{0}{\mathbf{T}}^{dr}_{1:2},\mathcal{Z}=\{{\mathbf{z}}_{1:2N}\}$: DR poses, side-scan measurements; \\
$n_{max},n_{sub}$: max iterations, subset size;
\Ensure ${}^{0}{\mathbf{T}}^{*}_{1:2}, \mathcal{X}^{*}_{sub}=\{{\mathbf{x}}_{1:n_{sub}}\}$;
\State $C^{*}_{p},C^{*}_{r},C^{*}_{o} \gets infinity$
\While{$n < n_{max}$}
\State $\mathcal{Z}_{sub} \gets RandomSample(\mathcal{Z},n_{sub})$
\State ${}^{0}{\mathbf{T}}_{1:2},\mathcal{X}_{sub},C_{o} \gets PoseEstimation({}^{0}{\mathbf{T}}^{dr}_{1:2},\mathcal{Z}_{sub})$
\State $C_{p},C_{r} \gets HypothesisCheck({}^{0}{\mathbf{T}}_{1:2},\mathcal{Z}\setminus\mathcal{Z}_{sub})$
\If{($C_{p}^{*}>C_{p}$ and $C_{r}^{*}>C_{r}$ and $C_{o}^{*}>C_{o}$)}
    \State $C^{*}_{p} \gets C_{p},\ C^{*}_{r} \gets C_{r},\ C^{*}_{o} \gets C_{o}$
    \State ${}^{0}{\mathbf{T}}^{*}_{1} \gets {}^{0}{\mathbf{T}}_{1},\ {}^{0}{\mathbf{T}}^{*}_{2} \gets {}^{0}{\mathbf{T}}_{2},\ \mathcal{X}^{*}_{sub} \gets \mathcal{X}_{sub}$
\EndIf
\State $n$++
\EndWhile
\State \Return ${}^{0}{\mathbf{T}}^{*}_{1:2}, \mathcal{X}^{*}_{sub}$
\end{algorithmic}
\end{algorithm}

The core idea of this pipeline is the consideration of both measurement cost ($\it{plane}$ $\it{cost}$ $C_{p}$ and $\it{range}$ $\it{cost}$ $C_{r}$) in the unsampled data, and $\it{optimization}$ $\it{cost}$ $C_{o}$ in the sampled data. The former measures how well the hypothesis model fits the unsampled data, while the latter indicates how well the optimization converges using the sampled data. We found experimentally that only using the measurement cost to determine the best model could be biased in many cases, while combining with optimization cost helps mitigate this issue by avoiding bad convergence. Therefore, we only update the current best model when both the costs are decreased. In additions, the final cost values also serve as indicators against empirical-based thresholds (see Section~\ref{sec:experi}-4), to filter out the poorly-estimated relative pose from being used as loop closure edge in pose graph optimization.




\subsection{Pose Graph Optimization}

We solve the SLAM problem through pose graph optimization using a factor graph formulation, as demonstrated in Fig.~\ref{fig:factorgraph} centre. Specifically, two types of measurements are considered in the problem: the odometry measurements and the loop-closure measurements. The dead-reckoning system provides a smooth yet drifted AUV poses, which can be used to form odometry measurements between consecutive poses as a chain in the graph. The odometry measurement error ${\mathbf{e}}_{n}({}^{o}{\mathbf{T}}_{n},{}^{o}{\mathbf{T}}_{n+1})$ is defined as:
\begin{eqnarray}
{\mathbf{e}}_{n}({}^{o}{\mathbf{T}}_{n},{}^{o}{\mathbf{T}}_{n+1}) = {}^{o}{\mathbf{T}}_{n}^{-1}\cdot{}^{o}{\mathbf{T}}_{n+1} - {}^{n}{\mathbf{T}}^{odo}_{n+1} + {\mathbf{\eta}},
\label{eq:odometry}
\end{eqnarray}
\noindent where ${}^{n}{\mathbf{T}}^{odo}_{n+1}$ is the odometry measurement obtained from dead-reckoning data. The estimated relative pose between the corresponding $i$ and $j$ subframes is served as a loop-closure measurement ${}^{i}{\mathbf{T}}^{lc}_{j}$ , such that the measurement error ${\mathbf{e}}_{m}({}^{o}{\mathbf{T}}_i,{}^{o}{\mathbf{T}}_{j})$ is denoted as:
\begin{eqnarray}
{\mathbf{e}}_{m}({}^{o}{\mathbf{T}}_i,{}^{o}{\mathbf{T}}_{j}) = {}^{o}{\mathbf{T}}_{i}^{-1}\cdot{}^{o}{\mathbf{T}}_{j} - {}^{i}{\mathbf{T}}^{lc}_{j} + {\mathbf{\eta}}.
\label{eq:odometry_and_lc}
\end{eqnarray}
Given that all the measurements follow Gaussian distributions, the AUV poses can be computed by minimizing the minus log of least squares cost as:
\begin{eqnarray}
\mathcal{C}({}^{o}{\mathbf{T}}_{1:N+1})= 
\frac{1}{2} (\sum^{N}_{n=1}[{\mathbf{e}}_{n}\Sigma{}^{-1}_{n}{\mathbf{e}}^{\top}_{n}] + \sum^{M}_{m=1}[{\mathbf{e}}_{m}\Sigma{}^{-1}_{m}{\mathbf{e}}^{\top}_{m}]),
\label{eq:globalcost}
\end{eqnarray}
where $N$ and $M$ is the number of odometry and loop-closure edges in the graph, respectively. $\Sigma{}_{n}$ is the odometry covariance that can be decided in a similar way as $\Sigma{}_{t}$, and $\Sigma{}_{m}$ is the covariance of loop-closure measurement that comes together with the estimated relative pose. Note that we omit the constant prior factor as shown in Fig.~\ref{fig:factorgraph} (centre) here to simplify the equation. We solve for Eq.~\ref{eq:globalcost} incrementally using iSAM2~\cite{kaess2012ijrr}, i.e., the whole pose graph is updated iteratively from the previous solution, when there are new measurements extracted from an input side-scan image and added to the graph. 
This iterative approach prevents drift from accumulating over a long trajectory. As such, wrong local minima caused by linearization far away from the optimal solutions can be avoided.

\subsection{Landmark Reconstruction}

After the pose graph optimization, a globally optimized AUV pose trajectory is obtained, which is used to reconstruct a quasi-dense map from the dense correspondences in all side-scan images. Specifically, we use Eq.~\ref{eq:costfunc} again, but fix the poses with the optimized ones and only optimize for the landmarks. Note that, as the dense correspondences are very noisy and full of outliers due to the edge issues discussed in Section~\ref{sec:ppg}, we discard part of the landmarks to be quasi-dense in quantity, via threshold on the plane and range costs, similar to loop closure edge filtering in Section~\ref{sec:method}-C.


\section{Experiments}
\label{sec:experi}

We evaluate our proposed SLAM framework in terms of pose trajectory and mapping. The evaluation is done on three different datasets that we name directly with their mission number: Mission1, Mission8 and Mission58.

\subsection{Experimental Setups}

\subsubsection{Dataset Description and Annotation}

\begin{figure}
 \centering
 \captionsetup{justification=centering}
 \frame{\includegraphics[width=1.\columnwidth]{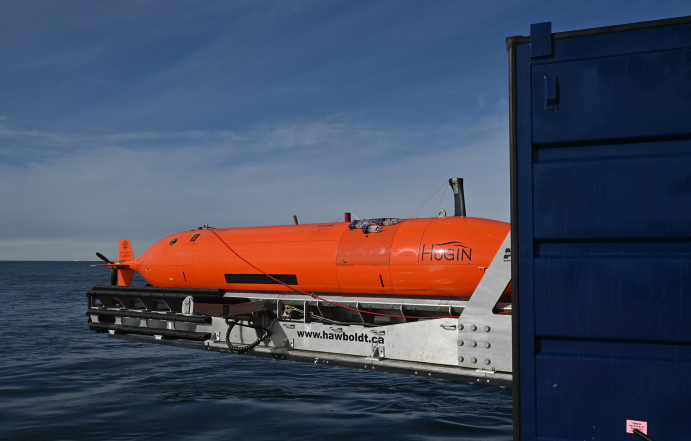}}
 \caption{The Hugin 3000 AUV used for collecting data in this work.}
 \label{fig:hugin}
 \vspace{-.4cm}
\end{figure}

All the side-scan sonar data tested in this work are collected by Gothenburg University’s Hugin AUV equipped with EdgeTech $2205$ side-scan sonar (see Fig.~\ref{fig:hugin}). The datasets were collecting approximately $4$ pings per second with the AUV speed at $2$m/s. The survey lines either following a lawn-mower or a loop pattern are roughly parallel to one another and have large overlap between each other. The seafloor of the surveyed areas are locally flat with gentle slope in altitude, and they contain lots of trawling marks. Details of the dataset and sonar characteristics can be found in Table~\ref{tab:datasets}.

In Mission1 dataset, we manually annotate sets of keypoint correspondences between each pair of the overlapped images and use them as ground truth reference for pose evaluation. The annotation process is conducted efficiently with the help of a $3$D mesh of the bathymetry that is constructed from MBES data. The $3$D mesh is used to find an initial guess of the potential corresponding keypoints when the annotator identifies a keypoint in the source image. Thus, given the proposed correspondence, the annotator only needs to inspect and confirm the correctness of the correspondence based on the image appearance. Using this method, we obtain about $250$$\sim$$500$ keypoint correspondences for each image pairs.

\begin{table}
  \centering
  \caption{Dataset and sonar characteristics.}
  \label{tab:datasets}
 \begin{tabular}{c|c|c|c}
  \toprule
                     &{Mission1}  &{Mission8} &{Mission58}    \cr
  \midrule
                     {Survey Pattern}     &Lawn-mower &Loop  &Lawn-mower       \cr
  \midrule
                     {Survey Area}      &$\sim200$x$800$m  &$\sim450$x$800$m &$\sim800$x$800$m      \cr
  \midrule
                     {No. of Lines}           &$5$  &$8$ &$14$        \cr
  \midrule
                     {Bins per Ping}           &\multicolumn{2}{|c|}{$\sim1200$} &$\sim2040$        \cr
  \midrule
                     {Total Pings}           &$9608$  &$18000$ &$24209$        \cr
  \midrule
                     {Mean Altitude}             &\multicolumn{3}{|c}{$\sim18$m}      \cr
  \midrule
                     {Max Range}    &\multicolumn{2}{|c|}{$170$m} &$80$m       \cr
  \midrule
                     {Frequency}        &\multicolumn{2}{|c|}{$410$KHz} &$850$KHz      \cr
  \bottomrule
 \end{tabular}
\end{table}

\begin{figure*}
 \centering
 \includegraphics[width=2.0\columnwidth]{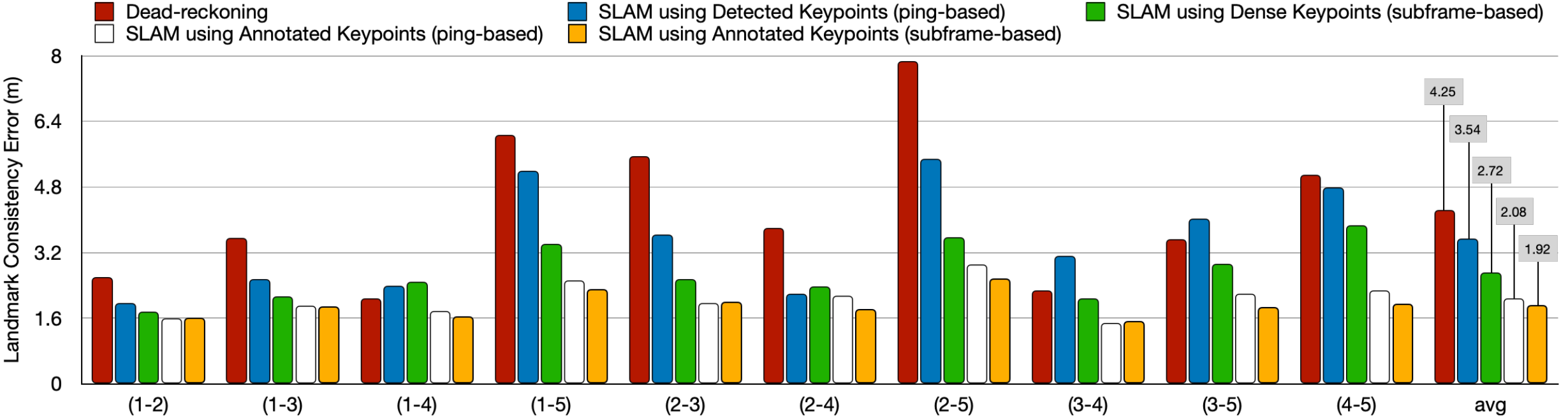}
 \caption{Landmark consistency errors (LCE) of the Mission1 dataset, which are averaged on all the annotated keypoint correspondences between each SSS image pair and over all pairs. Note that the X axis represents the index of each image pair. Ping-based SLAM results (blue and white) are generated by our previous method~\cite{zhang2023ietrsn}, while subframe-based SLAM results (green and orange) are generated by the proposed method in this paper.}
 \label{fig:lmc_errors}
\end{figure*}

\subsubsection{Evaluation Metrics}

\paragraph{Pose Metric} Due to lack of ground truth robot pose trajectory, which is often difficult to obtain in underwater, it is hard to evaluate pose trajectory with metric~\cite{sturm2012iros} that is commonly used for on-land or aerial robots. To mitigate such problem, we propose to compute the \textit{landmark consistency error} (LCE) using annotated keypoint correspondences and the bathymetry mesh from MBES data, considering the fact that a landmark observed from different ping pose should have a unique global position in ideal scenario. In particular, for each pair of keypoint correspondence, we project both keypoints onto the mesh by ray-casting and find their intersections with the mesh~\cite{bore2022joe}, respectively. Then, the Euclidean distance between both intersected landmarks is used as the error metric. The smaller this consistency error, the more accurate the ping poses. Note that this metric requires manually annotated keypoints, which is only available for Mission1.

\paragraph{Map Metric} To evaluate the bathymetry reconstruction result, one common way is to compute the heightmap at a fixed resolution, then compute the \textit{mean absolute error} (MAE) against a reference heightmap of MBES mesh~\cite{xie2022joe}\cite{xie2023joe}. However, this also requires ground truth pose to guarantee the high precision of MBES data. To address this, we propose to evaluate the reconstructed map line-by-line in the sensor frame coordinate. Specifically, we project the estimated landmarks in the global frame coordinate to each survey line in the sonar sensor frame coordinate, and align them with the raw multi-beam point cloud using known sensor offset. Then the MAE between the heightmaps of estimated landmarks and raw multi-beam point cloud is calculated as the error metric, which is free from the influence of pose.

\subsubsection{Baselines}

We use navigation data from the inbuilt dead-reckoning system of Hugin as baseline reference for comparison. The dead-reckoning solution embedded in the Hugin AUV is a high accuracy Doppler Velocity Log (DVL) aided Inertial Navigation System (INS) that can integrate various forms of positioning measurements from Inertial Measurement Unit (IMU) in $1$nmi/h class, DVL, compass and pressure aiding sensor, etc., in an error-state Kalman filter and smoothing algorithm to estimate position, velocity and attitude. The overall accuracy would be around $0.08\%$\footnote{\url{https://www.gu.se/en/skagerak/auv-autonomous-underwater-vehicle}} of the distance travelled. More specific details can be found in~\cite{jalving2003oceans}\cite{gade2003navlab}.

We also compare the proposed method with our previous work~\cite{zhang2023ietrsn}, a sparse ping-based SLAM framework that directly estimate the relative pose of associated pings using their keypoint correspondence without robust consideration.

\subsubsection{Implementation Details}
For all experiments we use the same parameters. In particular, the size (ping number) of each subframe is $200$, and the iteration number $n_{iter}=2$. For dense matching, $p_{size} = 13$, $n_{max} = 10$, $o_{max}=5$. For robust pose estimation, $\sigma_{r} = 0.1$, $\alpha = 0.1$, $n_{max} = 200$, $n_{sub}=6$. For landmark reconstruction (loop-closure edge filtering), the range and plane thresholds are $r_{thr} = 0.1\,(0.3)$ and $p_{thr} = 0.3\,(0.5)$. For other parameters that are not discussed in the paper, please refer to our source code. 

\subsection{Results}

\subsubsection{Pose Trajectory}

As shown in Fig.~\ref{fig:lmc_errors}, the accuracy of pose estimation is evaluated on Mission1 dataset by treating the results of running SLAM with annotated keypoints as ground truth references (white and orange). Overall, our proposed method (green) achieve consistent improvement against both the dead-reckoning (red) and ping-based SLAM (blue) baselines, with an average error reduction by up to $1.53$m ($46\%$). When comparing only between the ground truth references, interestingly, our subframe-based method (orange) with much less loop-closing constraints still slightly outperforms the ping-based baseline (white), which uses every two-ping measurement to estimate pose as loop-closing constraint. This suggests that our subframe-based method can robustly estimate highly accurate relative pose from noisy and outlier-prone measurements.

\begin{table}
  \centering
  \setlength{\tabcolsep}{2.4pt}
  \caption{Comparison of ATE between Dead-reckoning and SLAM estimations on Mission1. Bold number indicates best performance.}
  \label{tab:rmse}
 \begin{tabular}{cccc}
  \toprule
                     &Dead-reckoning    &Ping-based SLAM   &Subframe-based SLAM     \cr
  \midrule
  RMSE (m)        &$3.6583$     &$3.2742$   &$\mathbf{2.0749}$      \cr
  \bottomrule
 \end{tabular}
\end{table}

\begin{figure}
 \centering
 \includegraphics[width=1.\columnwidth]{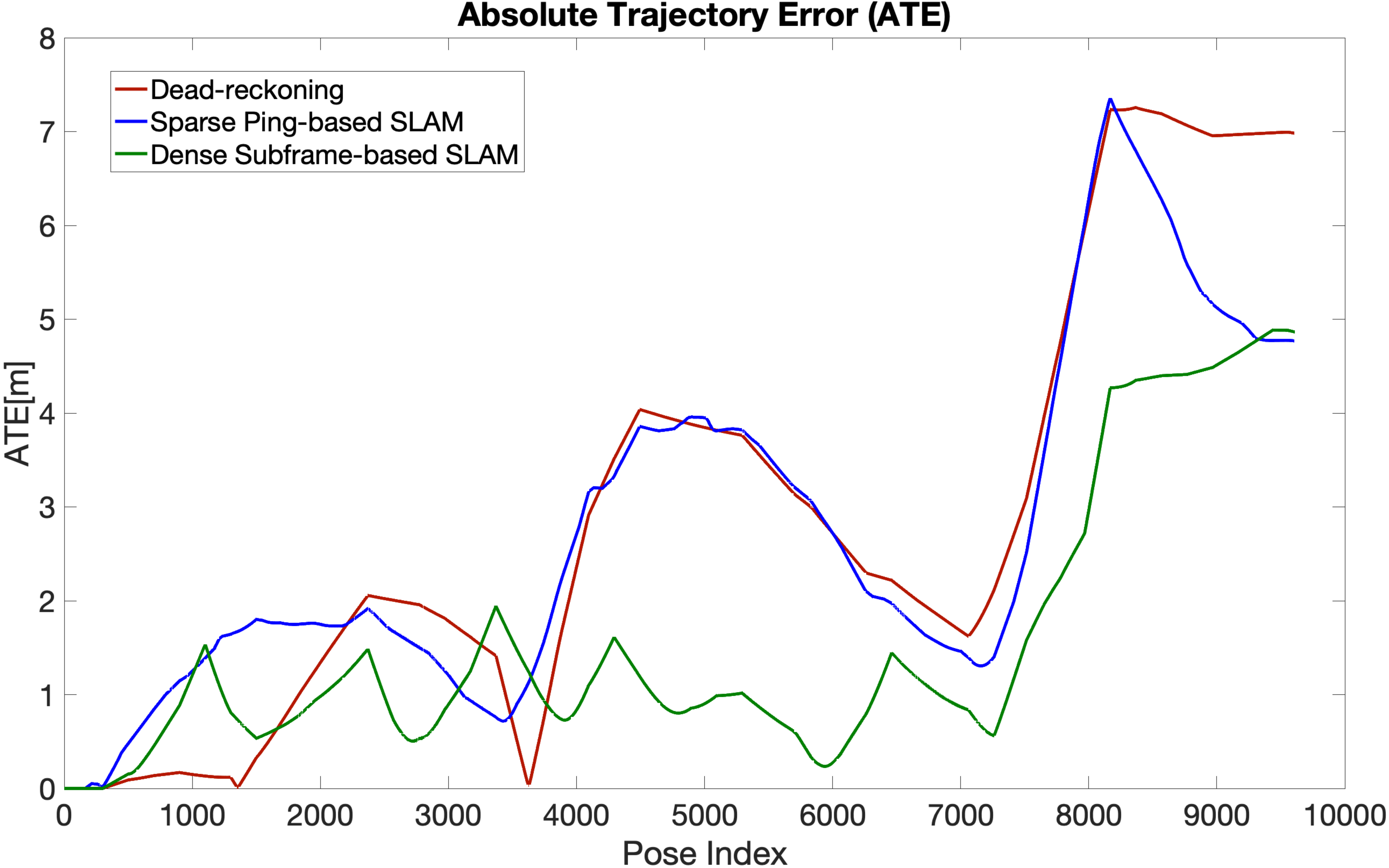}
 \caption{Comparison of absolute trajectory error (ATE) among dead-reckoning trajectory, estimations of the sparse ping-based SLAM baseline~\cite{zhang2023ietrsn} and our proposed dense subframe-based SLAM, on Mission1 dataset.}
 \label{fig:ate}
\end{figure}

\begin{figure}
 \centering
 \includegraphics[width=1.\columnwidth]{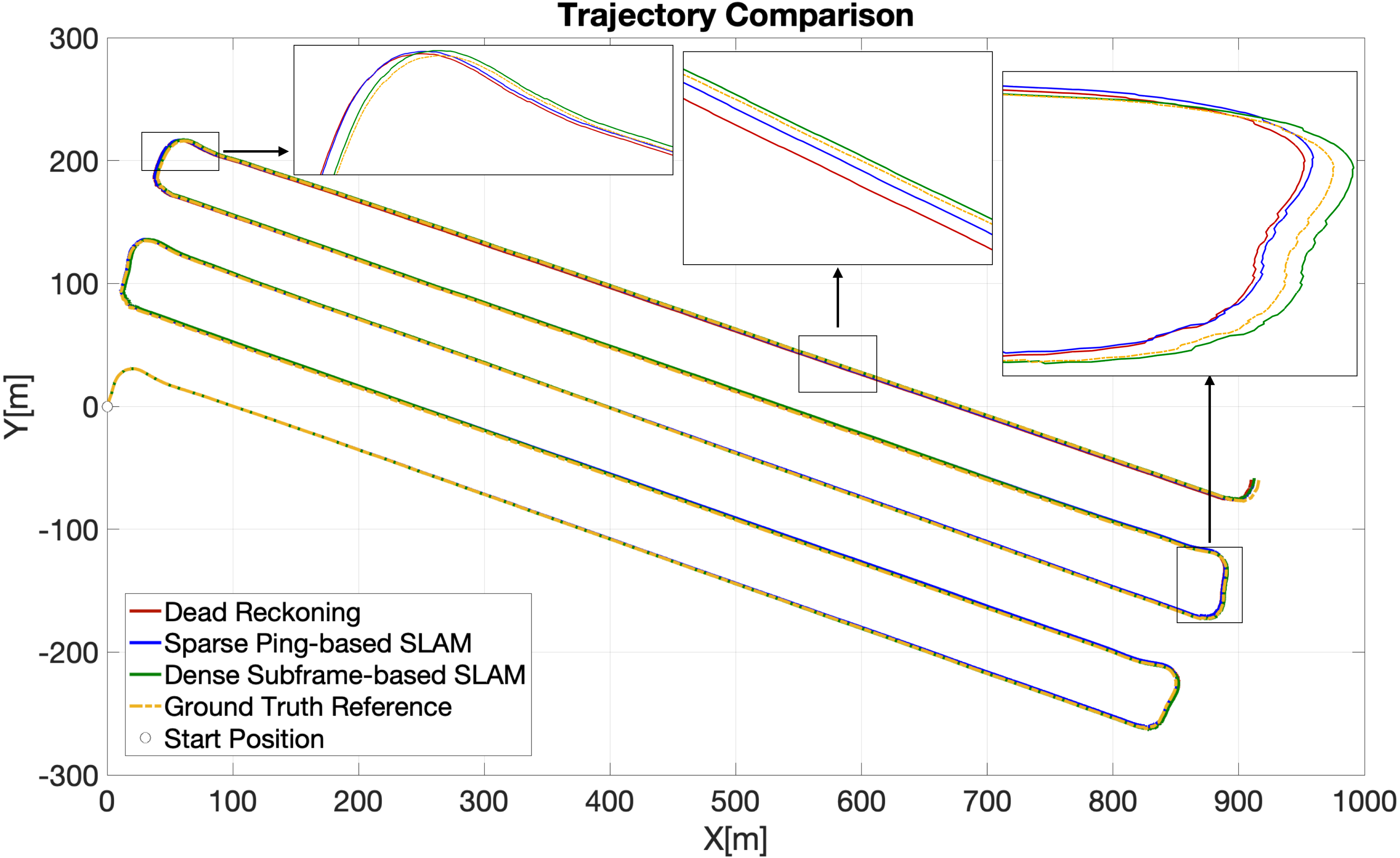}
 \caption{Comparison of Mission1 trajectories of the SLAM estimations and dead-reckoning data, with three zoom-in regions for better viewing.}
 \label{fig:auv_trj}
\end{figure}

\begin{figure*}
 \centering
 \includegraphics[width=2.0\columnwidth]{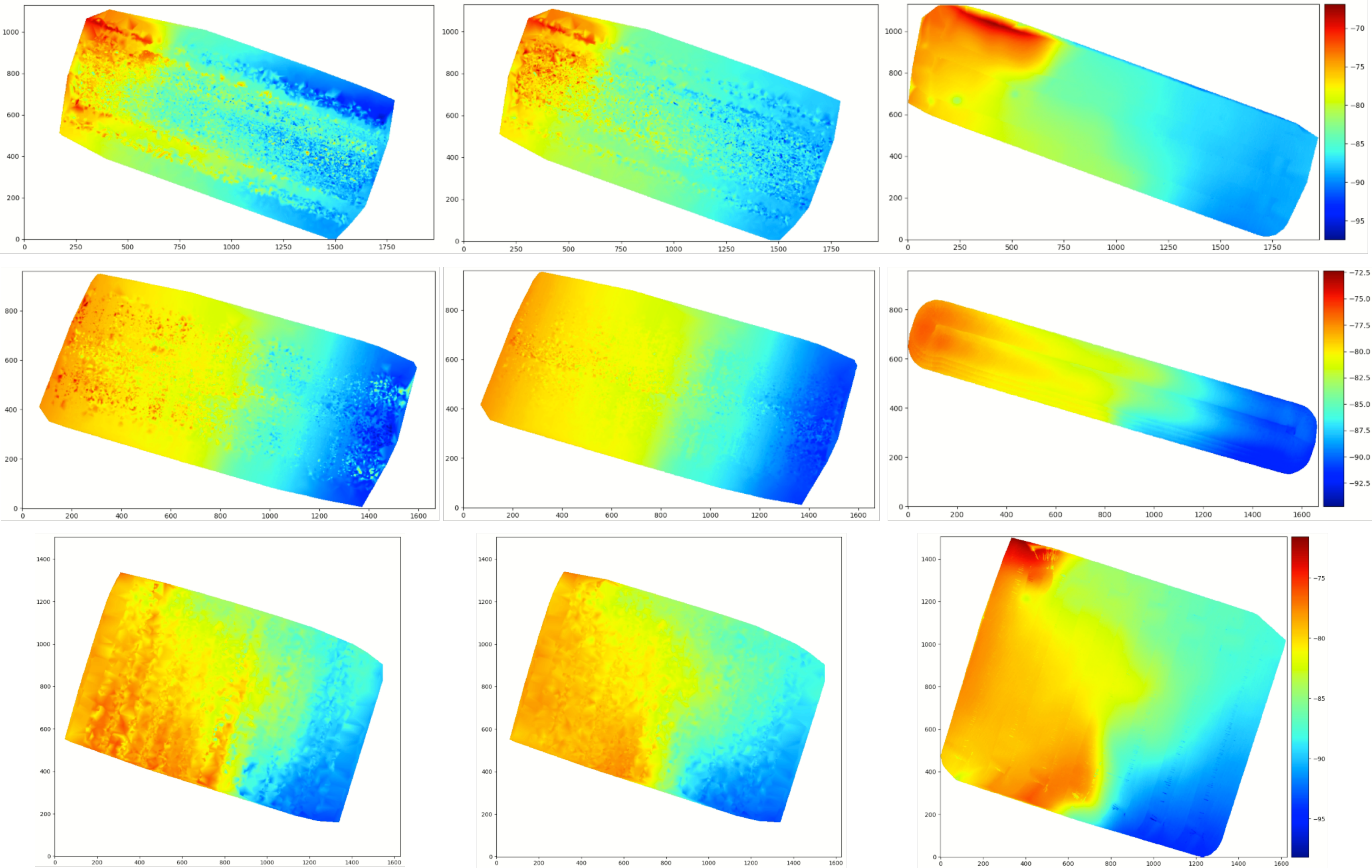}
 \caption{Heightmaps of the Mission1 (top row), Mission8 (centre row), and Mission58 (bottom row), generated by reconstructed bathymetry using dead-reckoning data (left column), using our proposed SLAM method (centre column) and using MBES mesh (right column).}
 \label{fig:heightmap}
\end{figure*}

We then compute the \textit{absolute trajectory error} (ATE)~\cite{sturm2012iros} to evaluate absolute pose consistency, by treating the trajectory generated by running our proposed SLAM with annotated keypoints (best performance in Fig.~\ref{fig:lmc_errors}) as ``ground truth'' trajectory. Fig.~\ref{fig:ate} demonstrates the error comparison of the whole trajectory. Although our estimated trajectory fluctuates slightly in the beginning, the ATE is reduced significantly after $4$k pings. As can be seen in Fig.~\ref{fig:auv_trj} the zoom-in windows, the estimated trajectory is closer to the ``ground truth'' trajectory, with about $43\%$ improvement over dead-reckoning in root mean squared error (RMSE), see Table~\ref{tab:rmse}. Interestingly, we can notice that the trajectory error with dead-reckoning does not always grow linearly over time, as Hugin running a lawnmower pattern could cancel out the drift growth obtained from body-fixed velocity and heading errors~\cite{jalving2003toolbox}. Similar effect can be observed in the landmark consistency error with dead-reckoning in Fig.~\ref{fig:lmc_errors}.

\subsubsection{Bathymetry Reconstruction}

We compare the quantitative results of our reconstructed bathymetry against the one reconstructed using dead-reckoning data on the three datasets, as demonstrated in Table~\ref{tab:heightmap_mae}. In particular, our proposed method consistently outperforms the baseline on Mission1 and Mission8 datasets, with an average improvement of $0.4$m ($28\%$) and $0.18$m ($22\%$), respectively. Our results of  Mission58 datasets are slightly better on average and with smaller variance than those of the baseline, though we have higher errors in some cases. We believe this is due to the side-scan images in this dataset having very low coverage, which leads to our method being only able to capture loop-closing constraints between adjacent survey lines. In this case, a global consistent pose trajectory cannot be achieved.

\begin{table}
  \centering
  \caption{Heightmap absolute mean errors (MAE in meters) of the three tested datasets. Bold values indicate best performance.}
  \label{tab:heightmap_mae}
 \begin{tabular}{c|cc|cc|cc}
  \toprule
  &\multicolumn{2}{c|}{Mission1}   &\multicolumn{2}{c|}{Mission8}       &\multicolumn{2}{c}{Mission58}       \cr
  \midrule
 Img. ID     &DR  &EST  &DR  &EST       &DR    &EST   \cr
  \midrule
  1  &1.22 &$\mathbf{1.01}$  &0.87  &$\mathbf{0.51}$  &0.32  &$\mathbf{0.26}$   \cr
  2  &1.52 &$\mathbf{1.12}$  &1.07  &$\mathbf{0.81}$  &0.63  &$\mathbf{0.58}$   \cr
  3  &1.27 &$\mathbf{0.98}$  &0.68  &$\mathbf{0.59}$  &0.48  &$\mathbf{0.48}$   \cr
  4  &1.29 &$\mathbf{1.02}$  &0.80  &$\mathbf{0.70}$  &$\mathbf{0.59}$  &0.71   \cr
  5  &2.04 &$\mathbf{1.21}$  &1.00  &$\mathbf{0.69}$  &0.63  &$\mathbf{0.47}$   \cr
  6  &    &$\mathbf{}$       &0.72  &$\mathbf{0.64}$  &0.70  &$\mathbf{0.59}$   \cr
  7     &    &       &0.97  &$\mathbf{0.81}$        &1.01  &$\mathbf{0.66}$   \cr
  8     &    &       &0.69  &$\mathbf{0.62}$        &0.82  &$\mathbf{0.67}$   \cr
  9     &    &       &  &                           &$\mathbf{0.49}$  &0.70   \cr
  10     &    &       &  &                          &$\mathbf{0.50}$  &0.54   \cr
  11     &    &       &  &                          &$\mathbf{0.53}$  &0.56   \cr
  12     &    &       &  &                          &0.61  &$\mathbf{0.52}$   \cr
  13     &    &       &  &                          &0.96  &$\mathbf{0.71}$   \cr
  14     &    &       &  &                          &$\mathbf{0.69}$  &0.86   \cr
  \midrule
  avg     &1.47 &$\mathbf{1.07}$  &0.85  &$\mathbf{0.67}$  &0.64  &$\mathbf{0.59}$   \cr
  \bottomrule
 \end{tabular}
\end{table}

We also show qualitatively the heightmaps generated from our reconstructed point clouds, the one reconstructed by dead-reckoning data, as well as the MBES mesh, see Fig.~\ref{fig:heightmap}. Specifically, the heightmaps obtained by our proposed method are very close to the multi-beam data, while have wider coverage (except for  Mission58). In particular, we can clearly see the misalignment between the multi-beam scans in Mission8, while our results are more globally consistent. Compared to the heightmaps generated by dead-reckoning data, our results are smoother with less noise.

\subsection{Discussion}

\subsubsection{Dense Matching Accuracy}

Given the annotated correspondences as ground truth baseline, we compute the \textit{end-point error} (EPE)~\cite{sun2014ijcv}, i.e., the pixel distance between the estimated and ground truth correspondence as metric for evaluation of side-scan image matching.

\begin{figure}[!htb]
 \centering
 \includegraphics[width=1.\columnwidth]{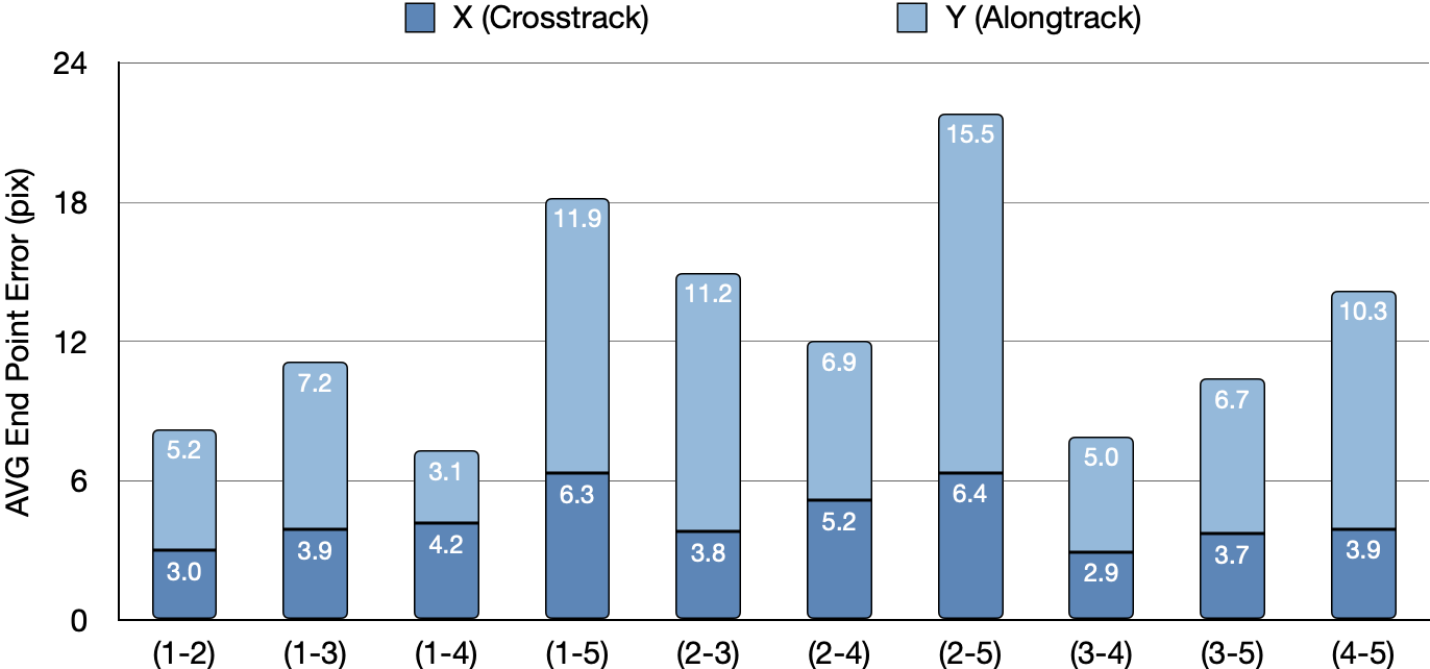}
 \caption{Averaged end-point errors (EPEs) of initial estimated correspondences between each SSS image pair of Mission1 dataset. Here X (image row) refers to AUV's crosstrack direction, and Y (column) the alongtrack direction.}
 \label{fig:epe_dr}
\end{figure}

Fig.~\ref{fig:epe_dr} illustrates the EPE of the estimated correspondences after initialization (Section~\ref{sec:ini}), i.e., only relying on geometric information (dead-reckoning) to find the closest match. We can see that the error in Y axis (longitudinal) is overall higher than that of X axis (lateral), which indicates the trajecory of AUV drifts more in the alongtrack direction. 

\begin{figure}[!htb]
 \centering
 \includegraphics[width=1.\columnwidth]{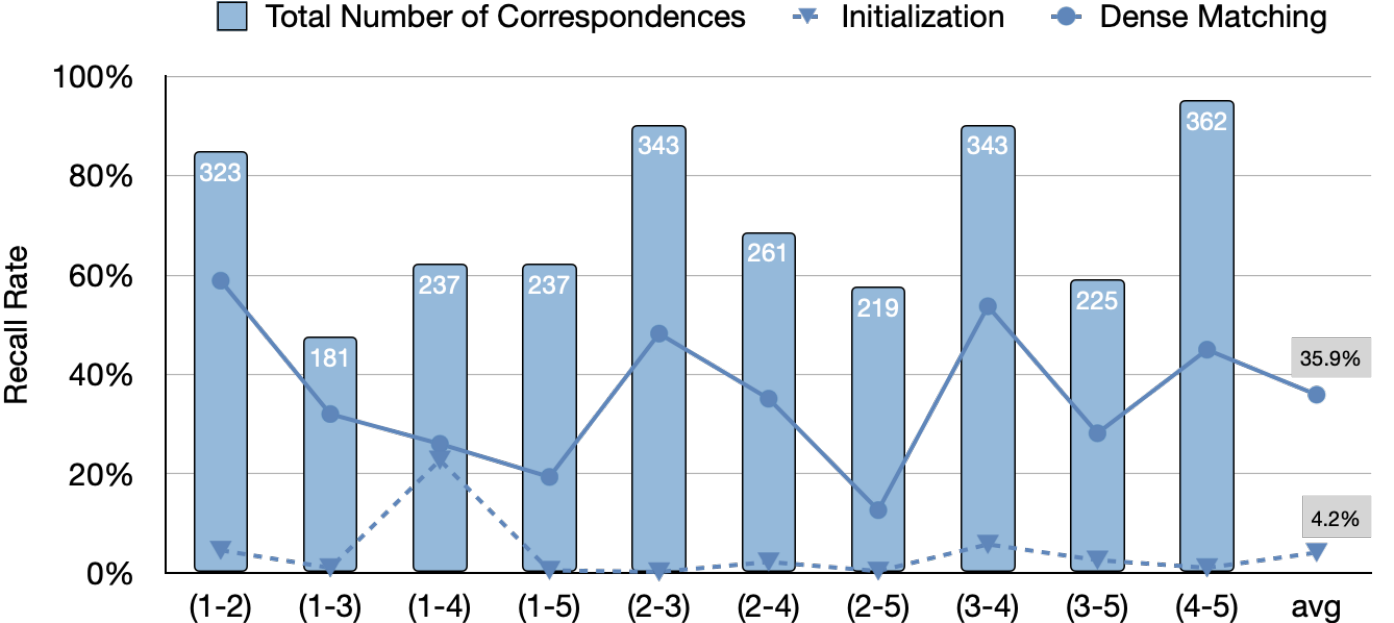}
 \caption{Recall rates of positive correspondences between each SSS image pair of Mission1 dataset, by initialization only and dense matching, respectively. Note that the total number of correspondences between each SSS image pair is shown with light grey bar for reference.}
 \label{fig:recall}
\end{figure}

Assuming that an estimated correspondence with $\leq2$ pixel error in both X and Y axis is a positive (good) correspondence, the recall rate after initialization is shown in Fig.~\ref{fig:recall} as the dashed line with triangle marks, which is fairly low with only $4.2\%$ on average, except for that of image pair ($1$-$4$), which could be due to the drifts being cancelled out in lawnmower pattern. After the full dense matching process, the recall rate is significantly increased with an improvement of $30\%$ on average, see the solid line with circle marks in Fig.~\ref{fig:recall}. Despite such improvement, the overall ratio of good correspondences is still very low for accurate pose estimation. Therefore, a robust estimation algorithm as our proposed method in Section~\ref{sec:spe} is essential.

\begin{figure}[!htb]
 \centering
 \includegraphics[width=1.\columnwidth]{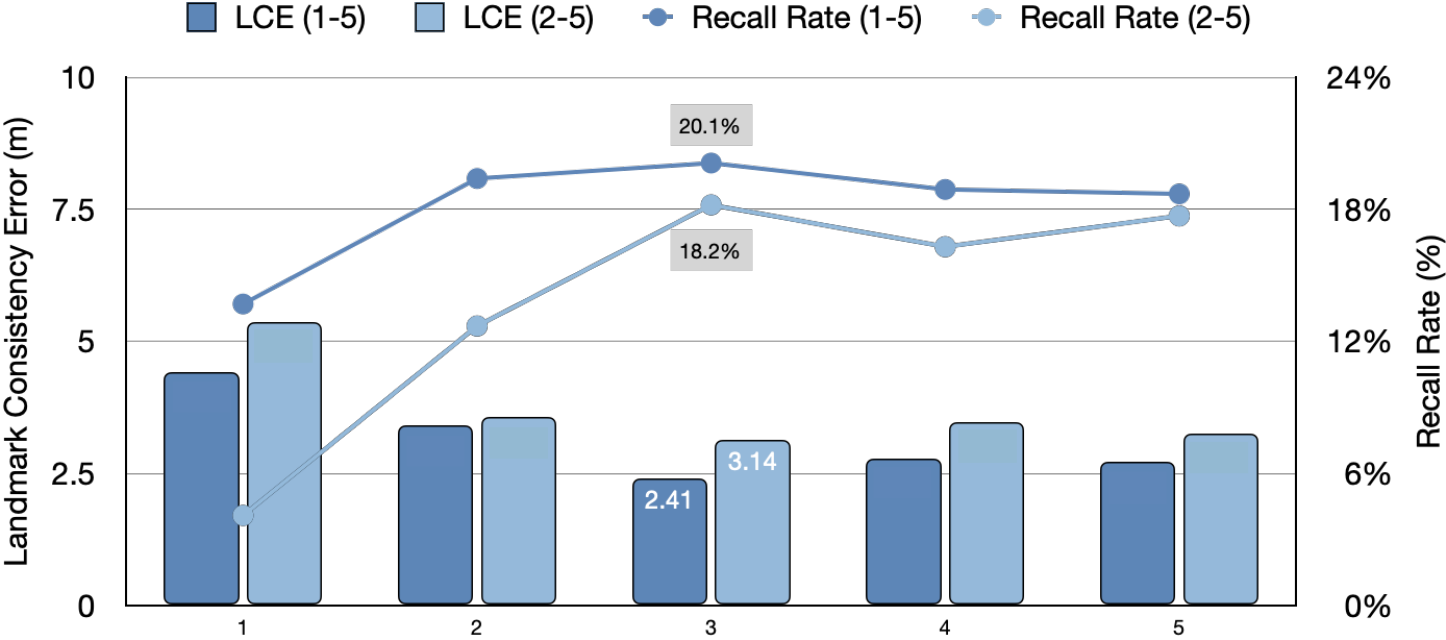}
 \caption{Influence of the iteration number $n_{iter}$ (X axis) on data association and pose optimization results of distant image pairs ($1$-$5$) and ($2$-$5$) from Mission1 dataset. The color bars represent the landmark consistency errors (LCE), which is corresponding to the left Y-axis. The lines with circle represent recall rate, which is corresponding to the right Y-axis. The best values are marked out for reference.}
 \label{fig:iterate}
\end{figure}

\subsubsection{Iterative Data Association with Optimized Poses}~\label{sec:dis2}
As described in Section~\ref{sec:iter}, the data association and pose optimization modules could benefit from each other in an iterative fashion. As illustrated in Fig.~\ref{fig:iterate}, the recall rates of both tested image pairs increase quickly in the first three iterations, and decrease slightly later. A similar trend can also be observed in their landmark consistency errors. From these results we believe that, by iterating between the data association and pose optimization process, the dense matching accuracy can be improved, and hence the improvement of pose estimation accordingly.

\begin{figure}[!htb]
 \centering
 \includegraphics[width=1.\columnwidth]{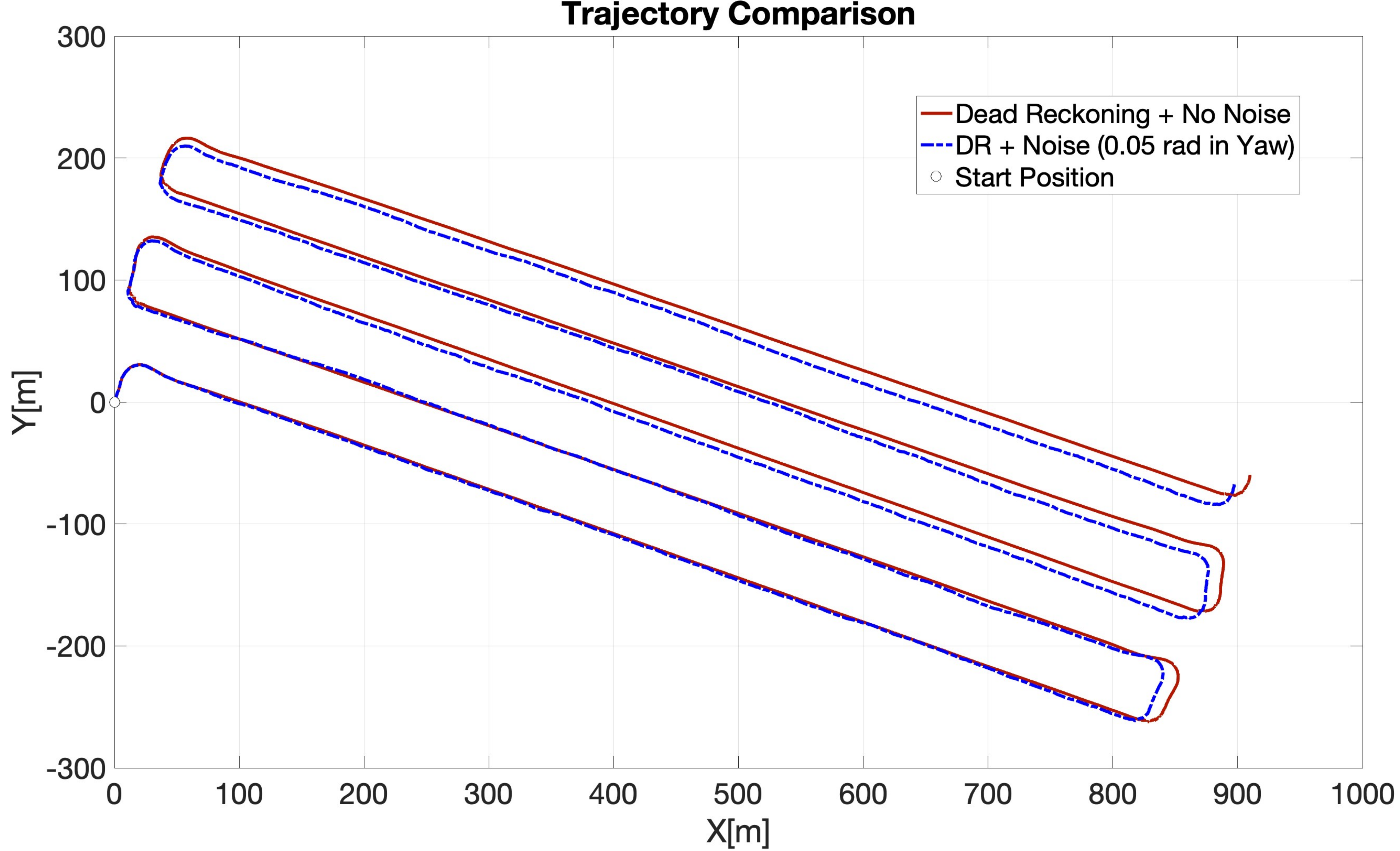}
 \caption{Dead-reckoning trajectories before and after adding Gaussian noise to the heading/yaw incrementally. Here the trajectory corrupted with noise level of $0.05$ rad ($\approx2.86$ deg), which is the maximum amount being tested in this paper, is demonstrated to better show the difference.}
 \label{fig:trjs_noise}
\end{figure}

\subsubsection{Method Feasibility against DR Accuracy }~\label{sec:dis3}
The proposed methods relies on reasonable odometry solutions (i.e., the Dead-reckoning system in this paper) to work properly. To evaluate the limit of our proposed method against the DR accuracy, we corrupt the Dead-reckoning trajectory by additionally adding different level of Gaussian noise to the heading (yaw) of the vehicle, to simulate larger accumulative drift in the DR estimates, as demonstrated in Fig.~\ref{fig:trjs_noise}. With the corrupted DR trajectory as input, we run our proposed method to get an optimized trajectory and compute the ATE against the "ground truth" reference trajectory, same as in Section~\ref{sec:experi}-B1.

\begin{figure}
 \centering
 \includegraphics[width=1.\columnwidth]{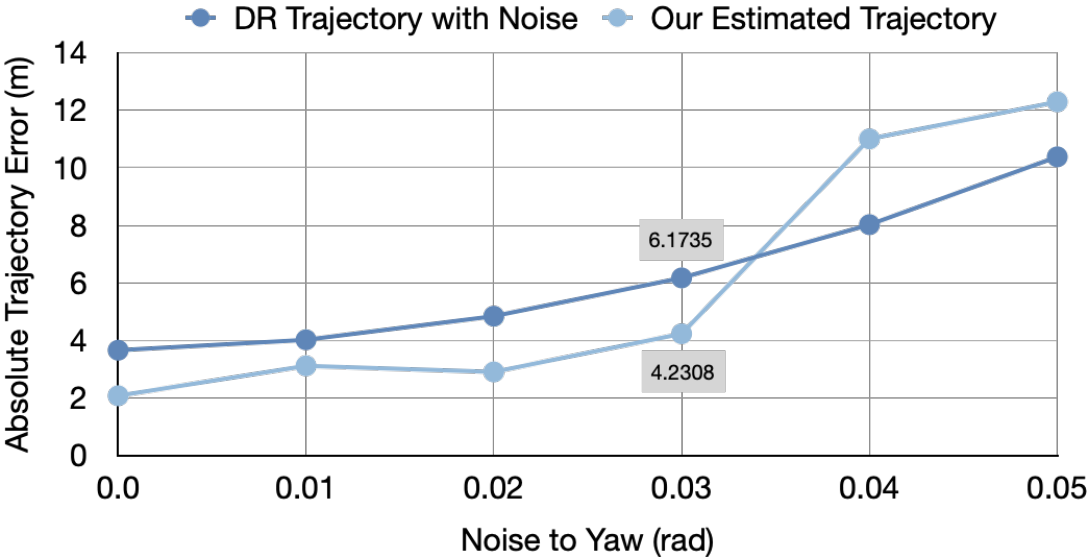}
 \caption{Illustration of Absolute Trajectory Error (ATE) between the DR trajectory with noise and our estimated trajectory, under different level of noise adding to the yaw angle.}
 \label{fig:noise_study}
\end{figure}

As shown in Fig.~\ref{fig:noise_study}, our method is able to work with a DR solution with up to $6.1$m absolute trajectory error, reducing drift by $31$\% to $4.2$m. Nevertheless, it fails to work properly after adding higher level of noises (after $0.04$ rad). The results become worse, mainly due to adding false loop-closure edges, which push certain parts of the trajectory away from the true position.

\section{Conclusion}
\label{sec:concl}

In this paper we present a dense subframe-based SLAM framework using side-scan sonar that is capable of improving the accuracy of AUV pose trajectory from dead-reckoning data, and reconstructing a quasi-dense bathymetry of the seabed. The proposed framework integrates an automatic dense matching method to effectively find dense correspondences between overlapping side-scan images, and utilize them to achieve robust and accurate estimation of poses between subframes, which are formulated as constraints to refine the pose trajectory through graph optimization. We carefully test and analyse our method on three different sets of real data collected by Hugin AUV, and demonstrate its effectiveness.

The proposed method is limited to reasonable DR solutions to work properly. If rather poor odometry is provided, our proposed system would either fail in finding valid loop closure edge (mistaken overlap check), or obtain invalid loop closure edge (bad data association). Another limitation is dependence on the assumption that the seafloor is relatively flat with gentle slope, primarily in consideration of canonical image transformation for dense matching, which may limits the applicable scenarios, e.g., regions with complex geometric information such as rocks, ridges, hills, etc. However, we may argue that the proposed method would be a good complementary solution for SLAM using multi-beam sensor, which usually requires the seabed to be geometric feature-rich for data association. But still and all, developing a robust and accurate data association algorithm for side-scan images without flat seafloor assumption and/or good odometry prior is a potential direction for future work. 

Another interesting direction could be to utilize the estimated quasi-dense landmarks as constraints for bathymetry reconstruction from side-scan images based on implicit neural representation, which can learn continuous, high-quality bathymetry using gradient-based optimization, while in our previous works~\cite{bore2022joe}\cite{xie2022joe} only sparse depth from altimeter readings are used. Denser bathymetric constraints from this work could potentially improve convergence speed and reconstruction quality, especially at regions that are far away from the AUV trajectory.  



%



\section*{Acknowledgment}

\noindent This work is supported by Stiftelsen f\"or Strategisk Forskning (SSF) through the Swedish Maritime Robotics Centre (SMaRC) (IRC15-0046) and the Wallenberg AI, Autonomous Systems and Software Program (WASP) funded by the Knut and Alice Wallenberg Foundation.

\ifCLASSOPTIONcaptionsoff
  \newpage
\fi


\bibliographystyle{IEEEtran}
\bibliography{refs/main}

%



%

\begin{IEEEbiography}[{\includegraphics[width=1in,height=1.25in,clip,keepaspectratio]{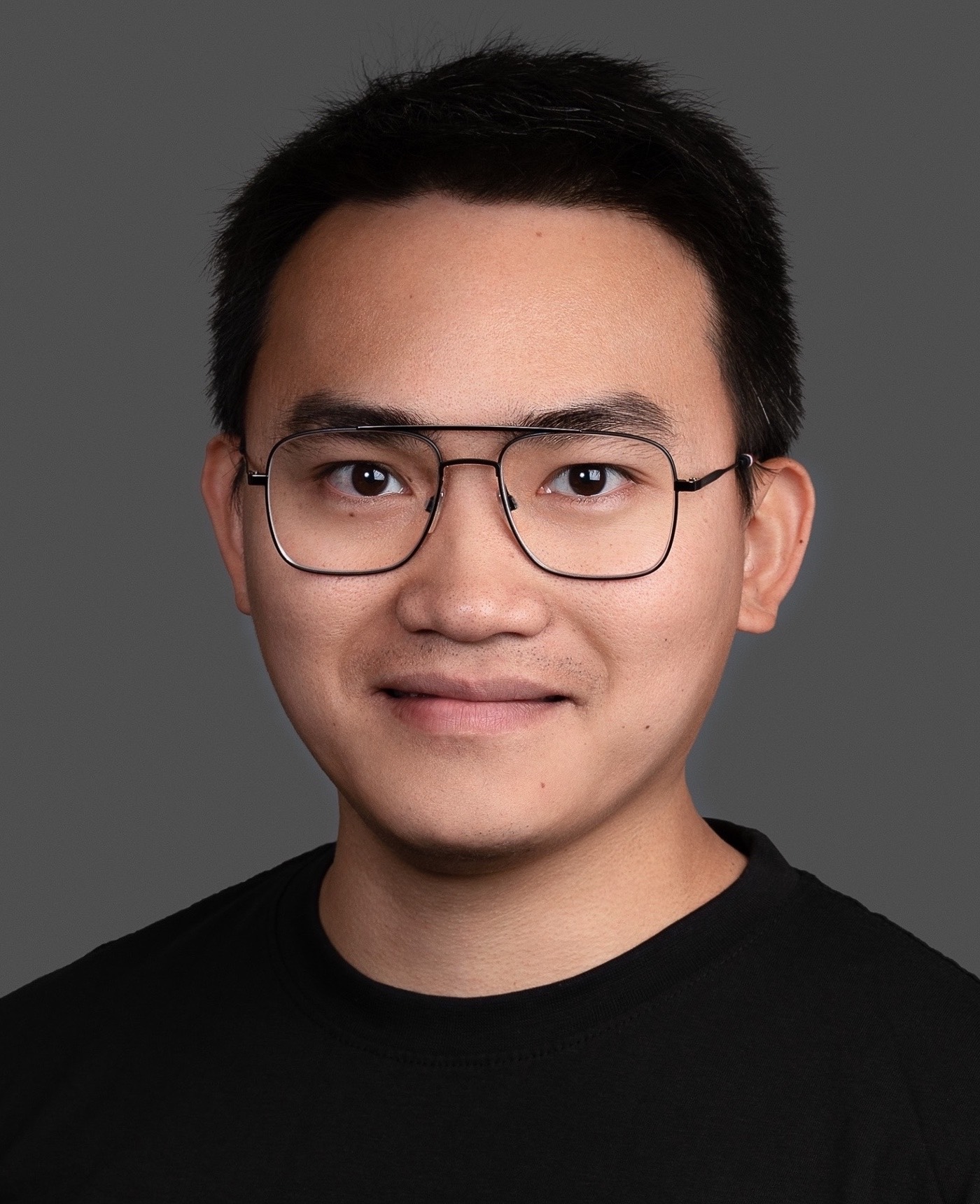}}]{Jun Zhang} 
received his Ph.D. degree from the Australian National University (ANU) in 2021, M.Sc.Eng. (2015) and B.Eng. (2012) degrees from Northwestern Polytechnical University (NPU). He is a research staff with the Institute of Computer Graphics and Vision, Graz Univeristy of Technology. His research interests include robot perception and 3D vision, in particular, simultaneously localization and mapping (SLAM) with acoustic and optical sensors.
\end{IEEEbiography}

\begin{IEEEbiography}
[{\includegraphics[width=1in,height=1.25in,clip,keepaspectratio]{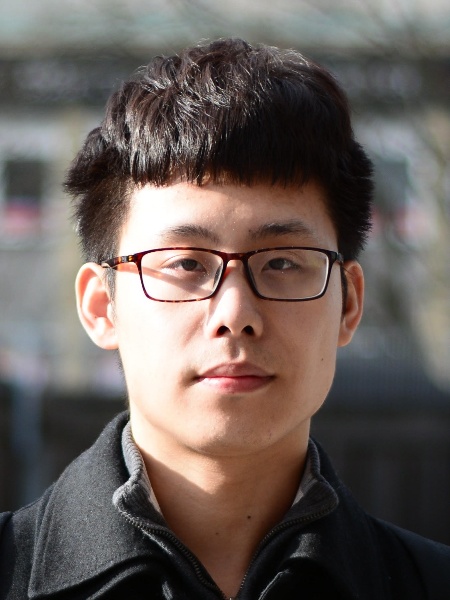}}]
{Yiping Xie} received the M.Sc. degree in computer science, and the Ph.D. degree in robotics from Royal Institute of Technology (KTH), Stockholm, Sweden, in 2019, and 2024, respectively. He is currently a researcher with the Swedish Maritime Robotics Centre (SMaRC) and Robotics Perception and Learning (RPL) Lab at KTH. His research interests include perception for underwater robots, bathymetric mapping and localization with sonars.
\end{IEEEbiography}

\begin{IEEEbiography}
[{\includegraphics[width=1in,height=1.25in,clip,keepaspectratio]{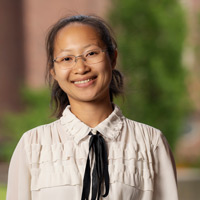}}]
{Li Ling} received the B.S. degree in computer science, and the M.Sc. degree in machine learning from Royal Institute of Technology (KTH), Stockholm, Sweden, in 2018 and 2021, respectively. She is currently a Ph.D. student with the Swedish Maritime Robotics Center (SMaRC) project in the division of Robotics, Perception and Learning (RPL) at KTH. Her research interests include perception for underwater robots and bathymetric reconstruction using sonars. 
\end{IEEEbiography}

\begin{IEEEbiography}[{\includegraphics[width=1in,height=1.25in,clip,keepaspectratio]{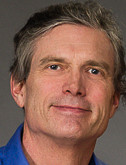}}]{John Folkesson}
 received the B.A. degree in physics from Queens College, City University of New York, New York, USA, in 1983, and the M.Sc. degree in computer science, and the Ph.D. degree in robotics from Royal Institute of Technology (KTH), Stockholm, Sweden, in 2001 and 2006, respectively. He is currently an Associate Professor of robotics with the Robotics, Perception and Learning Lab, Center for Autonomous Systems, KTH. His research interests include navigation, mapping, perception, and situation awareness for autonomous robots.
\end{IEEEbiography}




\end{document}